\documentclass[]{article}

\usepackage{jair, rawfonts, theapa}

\usepackage[ruled]{algorithm2e}
\usepackage{amsmath, amssymb}
\usepackage{amsfonts}
\usepackage{graphicx}
\usepackage{microtype}      % microtypography
\usepackage{nicefrac}       % compact symbols for 1/2, etc.
\usepackage{pgfplots}
\pgfplotsset{compat=1.13}
\usepackage{subcaption}
\usepackage{tikz}
\usepackage{url}            % simple URL typesetting
\renewcommand{\vec}[1]{\mathbf{#1}}
\newcommand{\mat}[1]{\mathbf{#1}}
\newcommand{\specialcell}[2][c]{\begin{tabular}[#1]{@{}l@{}}#2\end{tabular}}
\usetikzlibrary{shapes,trees,positioning, arrows,backgrounds, patterns}
\newcommand{\rpt}[2][1]{\foreach \n in {1,...,#1}{#2}}

\newcommand{\nnode}[4][0]{\node[rectangle,rounded corners,draw,fill=black!#1](#2)[#3]{$\rpt[#4]{\:\!\circ\:\!}$}}
\newcommand{\weight}[3]{\node[rectangle,rounded corners,dotted,draw,fill=black!5](#1)[#2]{#3}}
\newcommand{\inode}[3]{\nnode{#1}{#2}{2};\weight{#1W}{below =0cm of #1}{{\tt#3}}}
\newcommand{\rnn}[3]{\node(#1)[regular polygon,  regular polygon sides=3, rounded corners, draw, minimum width = 1.5cm, label=below:{{\tt#3}}][#2]{};}

\pgfarrowsdeclare{mytip}{mytip}{  
  \setlength{\arrowsize}{0.4pt}  
  \addtolength{\arrowsize}{.5\pgflinewidth}  
  \pgfarrowsrightextend{0}  
  \pgfarrowsleftextend{-5\arrowsize}  
}{  
  \setlength{\arrowsize}{0.3pt}  
  \addtolength{\arrowsize}{.5\pgflinewidth}  
  \pgfpathmoveto{\pgfpoint{-5\arrowsize}{4\arrowsize}}  
  \pgfpathlineto{\pgfpointorigin}  
  \pgfpathlineto{\pgfpoint{-5\arrowsize}{-4\arrowsize}}  
  \pgfusepathqstroke  
}

\jairheading{61}{2018}{907-926}{10/17}{04/18}

\ShortHeadings{How Neural Networks Process Hierarchical Structure}                 
{Hupkes, Veldhoen, \& Zuidema}                                           
\firstpageno{907} 

\title{Visualisation and `Diagnostic Classifiers' Reveal \\how Recurrent and Recursive Neural Networks \\Process Hierarchical Structure}

\author{\name Dieuwke Hupkes \email d.hupkes@uva.nl \\
        \name Sara Veldhoen \email sara.veldhoen@gmail.com \\
        \name Willem Zuidema \email zuidema@uva.nl \\
        \addr ILLC, University of Amsterdam\\
              P.O.Box 94242\\
              1090 CE  Amsterdam, Netherlands
}
\date{}

\begin{document}

\maketitle

%%%%%%%%%%%%%%%%%%%%%%%%%%%%%%%%%%%%%%%%%%%%%%%%%%%%%%%%%%%%%%%%%%%%%%
% Abstract

\begin{abstract}
 We investigate how neural networks can learn and process languages with hierarchical, compositional semantics.  
To this end, we define the artificial task of processing nested arithmetic expressions, and study whether different types of neural networks can learn to compute their meaning. 
We find that \textit{recursive} neural networks can implement a generalising solution to this problem, and we visualise this solution by breaking it up in three steps: project, sum and squash. 
As a next step, we investigate \textit{recurrent} neural networks, and show that a gated recurrent unit, that processes its input incrementally, also performs very well on this task: the network learns to predict the outcome of the arithmetic expressions with high accuracy, although performance deteriorates somewhat with increasing length. 
To develop an understanding of what the recurrent network encodes, visualisation techniques alone do not suffice.
Therefore, we develop an approach where we formulate and test multiple hypotheses on the information encoded and processed by the network.
For each hypothesis, we derive predictions about features of the hidden state representations at each time step, and train `\textit{diagnostic classifiers}' to test those predictions. 
Our results indicate that the networks follow a strategy similar to our hypothesised `cumulative strategy', which explains the high accuracy of the network on novel expressions, the generalisation to longer expressions than seen in training, and the mild deterioration with increasing length. This in turn shows that diagnostic classifiers can be a useful technique for opening up the black box of neural networks. We argue that diagnostic classification, unlike most visualisation techniques, does scale up from small networks in a toy domain, to larger and deeper recurrent networks dealing with real-life data, and may therefore contribute to a better understanding of the internal dynamics of current state-of-the-art models in natural language processing.
\end{abstract}

%%%%%%%%%%%%%%%%%%%%%%%%%%%%%%%%%%%%%%%%%%%%%%%%%%%%%%%%%%%%%%%%%%%%%%
% Introduction

\section{Introduction}

A key property of natural language is its hierarchical compositional semantics: the meaning of larger wholes depends not only on its words and subphrases, but also on the way they are combined.
Subphrases, in turn, can also be composed of smaller subphrases, resulting in sometimes quite complex hierarchical structures.
For example, consider the meaning of the noun phrase \emph{the scientist who wrote the natural language research paper}, which is a combination of the meanings of \emph{the scientist}, \emph{wrote} and \emph{natural language research paper}.
The meaning of the latter phrase -- a compound noun -- is a combination of the meanings of \emph{natural language} and \emph{research paper}, which are combinations of the meanings of the individual words.
Such hierarchical structures can be well represented by symbolic models \cite<most famously,>{chomsky1956three,montague1970universal}, but if and how they can be represented by (artificial) \textit{neural} models is an open question. 
This question has a long tradition in linguistics and computational neuroscience; recently, it has also received much attention from the field of natural language processing.

Although neural network models for language have been around since at least the 1980s, whether such models are fundamentally capable of capturing the type of generalisations that are required for computing the meaning of complex hierarchical structures is still largely unclear.
From a theoretical perspective, even the simplest recurrent neural networks are known to be Turing complete \cite{siegelmann1995computational} and many papers can be found that demonstrate the capability of (hand-crafted) recurrent networks to implement small context-free or even context sensitive languages \cite<e.g.>{gers2001lstm,rodriguez2001simple}.
However, both manually constructed and theoretical solutions rely on unrealistically high levels of precision and do therefore not scale up to real-life problems.
Furthermore, the theoretical existence of solutions is not of much help if they are intractable in the vast parameter spaces of neural networks, or if they cannot be learned from (a finite amount of) data.

On the practical side, recent advances have shown that both recursive and recurrent artificial neural networks perform extremely well on natural language processing tasks such as machine translation \cite<e.g.>{sutskever2014sequence,bahdanau2014neural}, sentiment analysis \cite<e.g.>{kalchbrenner2014convolutional,socher2013recursive,le2015forest}, semantic role labelling \cite<e.g.>{roth2016neural} and recognizing textual entailment \cite<e.g.>{bowman2016fast}. Although perfect performance on all of these tasks requires (implicit) knowledge of the underlying hierarchical structure and meaning of utterances of language, in practice it is unclear what sort of information the models are extracting from sentence representations to solve their task.
Hence, also these models do not yet provide a convincing demonstration that artificial neural networks are adequately computing the meaning of sentences with hierarchical structure, nor are they convenient for improving our understanding of the means by which hierarchy and compositionality can be implemented in large collections of interconnected artificial neurons.

A deeper insight into the internal dynamics of artificial neural networks while they process hierarchical structures could prove valuable both from a theoretical and a cognitive perspective, and is -- considering the difficulty of searching through the vast parameter space of high dimensional neural networks -- also interesting from an engineering point of view.
However, such insight is not easily acquired by studying natural language with all its complexities directly.
In this paper, we take a different approach.
To analyse the mechanism of processing hierarchical compositional structures isolated from all other aspects involved in processing language, we study an artificial language -- the language of arithmetics -- of which both sentences, phrases and lexical items have clearly (and symbolically) defined meanings. 
First we investigate if and how \textit{recursive} neural networks \cite{socher2010learning} can learn to correctly compute the meaning of sentences from this language, when their structure is given as part of the input.
We present a thorough analysis of how they process them, using visualisations of the individual operations conducted by the network.
We then extend our investigation to \textit{recurrent} neural networks \cite{cho2014properties,elman1990finding} which take a step forward to resembling the human brain, and constitute the current state-of-the-art in natural language processing.

In this paper, a revised and thoroughly extended version of the work presented in \citeA{veldhoen2016diagnostic}, we start with a definition of the arithmetic language that is the topic of our studies in Section~\ref{sec:language}; in Section~\ref{sec:models}, we define the models we are studying; Section~\ref{sec:training} contains a description of our training methods, as well as a brief evaluation of the performance of the models on the defined task.
The main contribution of this paper can be found in Section~\ref{sec:analysis}, where we present the analysis of the internal dynamics of the trained models.

The result of this study is a comprehensive analysis of the implementation of the composition function in the recursive neural network as well as an insightful interpretation of the symbolic strategy followed by the recurrent networks.
Furthermore, we present a novel method to analyse high dimensional recurrent networks when visualisation techniques fall short.
We use this method to gain more insight in our own trained networks, but also argue that it can be useful for analysis in many other cases where high dimensional recurrent neural networks are involved.

%%%%%%%%%%%%%%%%%%%%%%%%%%%%%%%%%%%%%%%%%%%%%%%%%%%%%%%%%%%%%%%%%%%%%%%%%%%%%%%%
% Arithmetic Language

\section{Arithmetic Language}\label{sec:language}

\begin{figure}[!ht]
    \centering
    \setlength\tabcolsep{4mm}
    \renewcommand{\arraystretch}{1.3}
    \begin{tabular}{lll}
        & \multicolumn{1}{c}{\textit{Form}} & \multicolumn{1}{c}{\textit{Meaning}}\\
        \hline
        vocab & \specialcell{$\{\texttt{-ten, -nine, \ldots, nine, ten,}$\\
        \hspace{3mm} $\texttt{plus, minus, left\_bracket, right\_bracket}\}$} & \specialcell{$\{\text{-10, -9, \ldots, 9, 10,}$\\\hspace{12.5mm} $\text{ +,  -,  (,  )}\}$} \\
        $\mathbf{L1}$       & $\left\{\texttt{-ten, -nine, \ldots, nine, ten}\right\}$ & $\left\{\text{-10, -9, \ldots, 9, 10}\right\}$ \\

        $\mathbf{Lk}$   & \specialcell{$\{\texttt{(}\;\mathbf{l_m}\;\texttt{op}\;\mathbf{l_n}\;\texttt{)}\,|\,\mathbf{l_m}\!\in\!\mathbf{Lm},\, \mathbf{l_n}\!\in\!\mathbf{Ln},\,\, \texttt{op}\in\{\texttt{plus},\, \texttt{minus}\},$\\
        \hspace{46mm} $m+n = k\}$} & $\langle l_m\rangle\, \langle\;\texttt{op}\;\rangle\,\langle\,l_m\rangle$\\
     \end{tabular}
     \caption{Formal description of the sentences $s$ in the arithmetic language and their meanings $\langle s \rangle$.}\label{fig:arithmetic_language}
\end{figure}

The vocabulary of the artificial language we consider consists of words for all integers in the range $\{-10,\ldots,10\}$, the operators {\tt plus} and {\tt -} and the brackets {\tt (} and {\tt )}. 
The grammatically correct phrases -- i.e.\ sequences of words -- in this \textit{arithmetic language} comprise all grammatically correct, fully bracketed arithmetic expressions that can be formed with these symbols.
The (compositional) meaning of an expression is the solution of the arithmetic expression that it represents.
For instance, the meaning of the phrase \texttt{( ten minus ( five plus three ))} is 2.
Troughout this paper, we will often abbreviate the full forms such as \texttt{left\_bracket five plus three right\_bracket} as \texttt{( 5 + 3 )}.

We refer to expressions and sets of expressions by using the number of numeral words they contain (see Figure~\ref{fig:arithmetic_language} for a formal definition).
For instance, $\mathbf{L5}$ refers to all expressions with exactly 5 numerals and $\mathbf{l_5}$ is an expression belonging to $\mathbf{L5}$.
Table~\ref{tab:language} contains some example sentences of the arithmetic language, along with the name of the subset they may appear in.

\begin{table}
    \centering
    \begin{tabular}{ll}
        \textbf{L1} & \texttt{one}, \hspace{5mm} \texttt{-three}, \hspace{5mm} \texttt{nine} \\
        \textbf{L2} & \texttt{(} \texttt{five} \texttt{plus} \texttt{three} \texttt{)} \\
        \textbf{L3} & \texttt{(} \texttt{(} \texttt{two} \texttt{minus} \texttt{-three} \texttt{)} \texttt{minus} \texttt{six} \texttt{)},\hspace{5mm} \texttt{(} \texttt{two} \texttt{minus} \texttt{(} \texttt{-three} \texttt{minus} \texttt{six} \texttt{)} \texttt{)} \\
                \textbf{L4} & \texttt{(} \texttt{(} \texttt{(} \texttt{-two} \texttt{minus} \texttt{seven} \texttt{)} \texttt{plus} \texttt{eight} ) \texttt{plus} \texttt{-ten}\texttt{)} \\
    \end{tabular}
    \caption{Sentences from different subsets of the arithmetic language. Both numerals, operators and brackets are treated as words; words are represented by $n$-dimensional numerical vectors.}\label{tab:language}
\end{table}

The arithmetic language is specifically chosen to allow us to study the mechanism of hierarchical compositionality in isolation, separate from other important aspects of natural language, such as structural and lexical ambiguity, irregular paradigms, multi-word units and idiomatic expressions.
Furthermore, the symbolic nature of the arithmetic language allows us to formulate strategies to compute the meaning of expressions, which can be used to aid analysis of the dynamics of the internal dynamics of a network processing sentences.
Before moving on to the description of our models and experiments, we will describe two possible strategies to incrementally solve arithmetic expressions.

 \begin{figure}[!h]
     \centering
     \begin{subfigure}[b]{0.49\linewidth}
 \input{figures/pseudo_recstrat}
 \caption{Recursive strategy}\label{recstrat}
 \end{subfigure}
 \begin{subfigure}[b]{0.49\linewidth}
 \input{figures/pseudo_immstrat}
 \caption{Cumulative strategy}\label{cumstrat}
 \end{subfigure}
 \caption{Different symbolic strategies for incrementally solving arithmetic expressions. The function {\tt apply}(\textit{mode}, \textit{result}, \textit{symbol}) applies the operator specified by \emph{mode} ($+,-$) to the two numbers specified by \emph{result} and \emph{symbol}.}\label{fig:pseudo}
 \end{figure}
 
\subsection{Recursive Strategy}
Perhaps the most obvious candidate for a symbolic strategy to compute the meaning of an arithmetic expression involves traversing through the expression, computing the outcome of all subtrees, until an outcome for the full tree is reached. 
 To do this in an incremental fashion, the intermediate result of the computation of the current subtree should be pushed onto a stack -- the \texttt{result\_stack} -- whenever a new, smaller subtree begins.
At that point, also the operator that will later be used to integrate the outcome of the newly started subtree with its parent, should be stored on a stack.
 Because this stack determines whether the procedure is in \emph{additive} or \emph{subtractive mode}, we call it the \texttt{mode\_stack}.
 Figure~\ref{recstrat} contains a procedural description of this strategy; for a worked out example, see the upper part of Figure~\ref{fig:processing}.
 
\subsection{Cumulative Strategy}
Alternatively, the meaning of a sentence from the arithmetic language can be computed by accumulating the numbers immediately at the moment they are encountered (see Figure~\ref{cumstrat} and the bottom part of Figure~\ref{fig:processing}).
 This means that at any point during the computation a prediction of the solution of the expression is maintained. 
 Consequently, this cumulative strategy does not require a stack with previous results, but it does require keeping track of previously seen operators to decide whether the next number should be added or subtracted when a bracket closes (in Figure~\ref{cumstrat} captured by the variable \texttt{mode}).

\subsection{Predictions Following from Strategies}
\begin{figure}[!ht]
     \centering
     \includegraphics[scale=0.60]{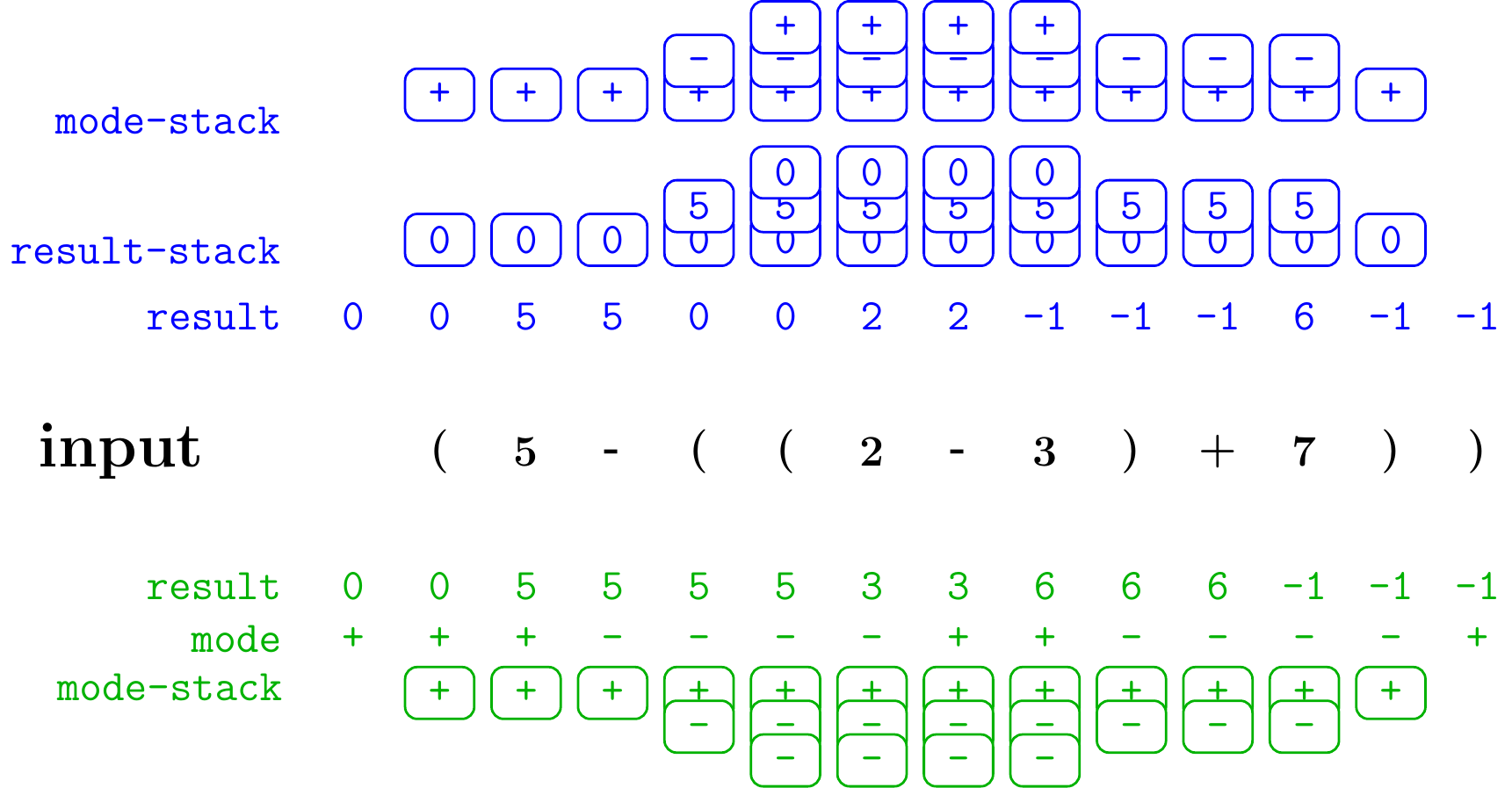}
     \caption{Different strategies to compute the meaning of a sentence of the arithmetic language. Top (in blue): \texttt{mode-stack}, \texttt{result-stack} and current \texttt{mode} of the recursive strategy; bottom (in green): cumulative \texttt{result}, \texttt{mode} and \texttt{mode-stack} of the cumulative strategy.}\label{fig:processing}
 \end{figure}

As illustrated in Figure~\ref{fig:processing}, the two strategies do not only differ with respect to the computations they are executing, but also require different memory contents.
 Both the cumulative and the recursive strategy require a stack to store encountered operators.
 Consider for instance computing the outcomes of the expressions \texttt{( 8 \textcolor{red}{-} ( ( 5 - 7 ) - 2 ) )} and \texttt{( 8 \textcolor{red}{+} ( ( 5 - 7 ) - 2 )}, for which information about the previous series of operators is required to understand whether the \texttt{2} should be subtracted or added to the subtotal.
 In addition, the recursive strategy requires storage of the previously computed outcomes of subtrees.
 These differences result in different predictions about the sensitivity of the network to noise on the stack, implementation of the operator or depth of an expression, and hence in different predictions about the difficulty of processing certain structures under memory limitations and noise.
 
%%%%%%%%%%%%%%%%%%%%%%%%%%%%%%%%%%%%%%%%%%%%%%%%%%%%%%%%%%%%%%%%%%%%%%%%%%%%%%%%
% Models

\section{Models}\label{sec:models}

We investigate two different types of artificial neural networks: \textit{recursive} neural networks (TreeRNNs) and \textit{recurrent} neural networks (RNNs).
The former process sequences recursively, following a given syntactic structure, whereas the latter process sequences sequentially, reading them one word at the time.

\subsection{TreeRNN}

The TreeRNN \cite{goller1996learning,socher2010learning} is a hybrid neural-symbolic model, in which a neural composition function is applied recursively, following a symbolic control structure.
The key component of the TreeRNN is the composition function, that computes a representation of the composition of two or more words or phrases.
In its most simple form this composition function is a single-layer feedforward network:

\begin{equation}
\vec{p}=\tanh(\mat{W}\cdot [\vec{x}_1;\dots;\vec{x}_n] +\vec{b}),\label{e:treeNN}
\end{equation}

\noindent where $\mat{W} \in \mathbb{R}^{d\times nd}$ and $\vec{b}\in \mathbb{R}^d$ are learned.
The input $[\vec{x_1};\dots;\vec{x_n}]$ is a concatenation of the input phrases or words $\{\vec{x}_1$,\dots$,\vec{x}_n\}$.
In case of the arithmetic language,  there are three input vectors $\vec{x}_1$, $\vec{x}_2$ and $\vec{x}_3$, representing a numeral, an operator and a numeral, respectively.

\begin{figure}[h]
    \centering
    \begin{subfigure}[b]{0.45\linewidth}
    \begin{tikzpicture}
\inode{p2}{}{$\mat{W}$};
\inode{p1}{below right =of p2W}{$\mat{W}$};
\draw[->] (p1) -> (p2W);
\inode{c4}{below left =of p2W}{five};
\inode{c3}{below =of p2W}{minus};
\draw[->] (c3) -> (p2W);
\draw[->] (c4) -> (p2W);
\inode{c1}{below left =of p1W}{two};
\draw[->] (c1) -> (p1W);
\inode{c2}{below  =of p1W}{plus};
\draw[->] (c2) -> (p1W);
\inode{c3}{below right=of p1W}{three};
\draw[->] (c3) -> (p1W);
\end{tikzpicture}
    \caption{}\label{f:mathTreeRNN}
    \end{subfigure}
    \begin{subfigure}[b]{0.54\linewidth}
    \centering
    \begin{tikzpicture}[>=latex,node distance = 0.5cm] 

\weight{Wc}{}{$W_{cl}$};
\nnode[20]{c}{above =0cm of Wc}{3};
\node(classes)[above of= c]{$<,=,>$};
\nnode[20]{q}{below = of Wc}{2};
\weight{Wq}{below=0cm of q}{$W_{comp}$};
\draw[->] (q) -> (Wc);

\rnn{t}{below left = 1cm and 0 cm  of Wq}{5+(2-3)};
\draw[->](t.90)->(Wq);

\rnn{t}{below right = 1cm and 0 cm  of Wq}{(4--8)-5};
\draw[->](t.90)->(Wq);
\end{tikzpicture}
\caption{} \label{fig:comparisonTraining} \end{subfigure}
    \caption{(a) A TreeRNN to compute the meaning of the sentence {\tt ( 5 - ( 2 + 3 ) )}. (b) Training setup of the TreeRNN\@. A comparison layer, on top of two TreeRNNs, is trained to predict whether left expression is smaller than, equal to or larger than the right one.}
\end{figure}
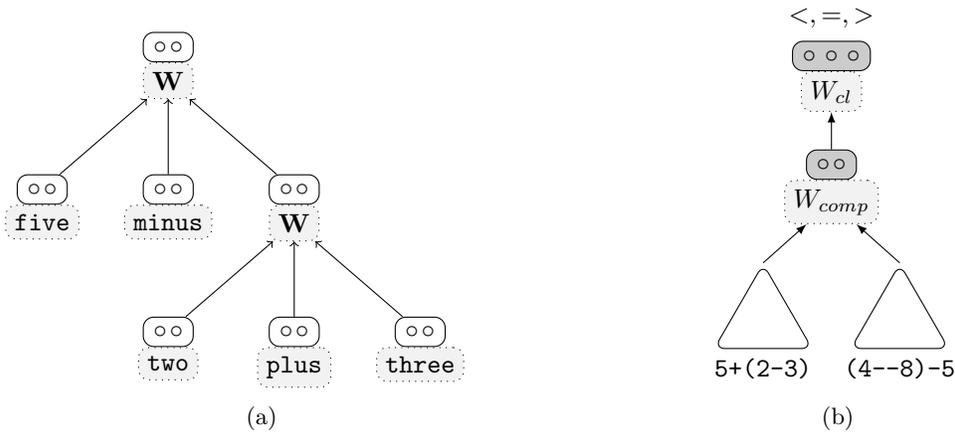

To compute the meaning of a sentence, the composition function is applied recursively, computing representations for all nodes in an externally provided phrase-structure tree in a bottom up fashion.
Thus, to process the tree in Figure~\ref{f:mathTreeRNN}, first the vectors for \texttt{two}, \texttt{plus}, and \texttt{three} are used, and Equation~\ref{e:treeNN} is used to compute a vector for their combination. 
This new vector is then, in turn, combined with vectors for \texttt{five} and \texttt{minus}, to compute the meaning for the entire expression using again Equation~\ref{e:treeNN}. 
Note that the vectors representing words and those representing combinations of words have the same dimensionality.

As the syntactic structure of sentences in the arithmetic language is included in their definition, -- and is thus unambiguous -- training a TreeRNN to compute their meanings allows us to study how a composition function can be implemented in a neural model.

\subsection{RNN}

In addition to TreeRNNs, we study the dynamics of their sequential counterparts: RNNs, which will allow us to more easily scale up our findings and techniques to current state-of-the-art models.
In particular, we consider two types of sequential networks: simple recurrent networks (SRNs), as originally introduced by \citeA{elman1990finding}, and their more recent extension Gated Recurrent Units (GRUs), defined by \citeauthor{cho2014properties}. 

\subsubsection{SRN}

\begin{figure}[!h]
    \begin{subfigure}[b]{0.49\linewidth}
        \centering
        \includegraphics[width=0.9\textwidth]{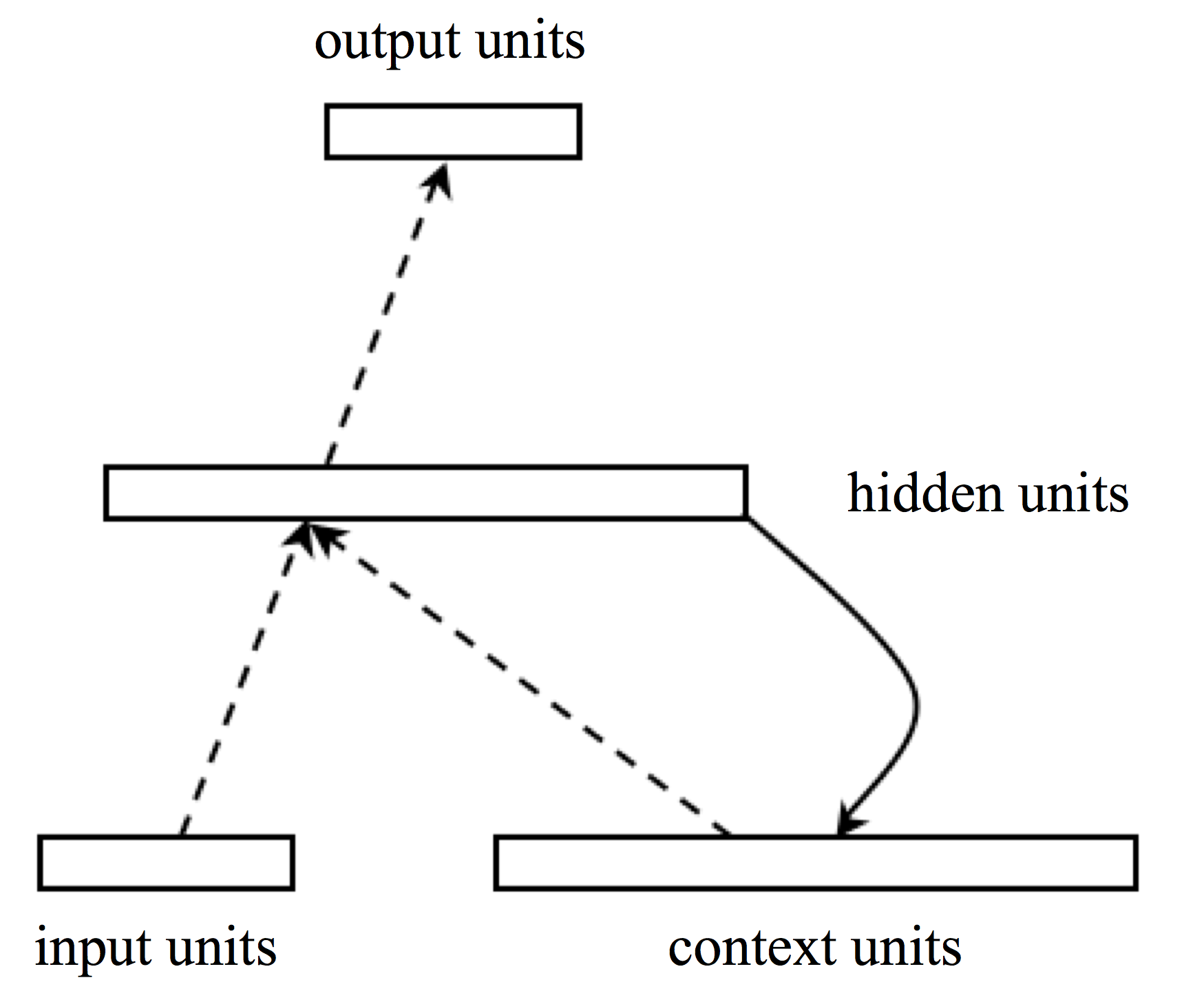}
        \caption{SRN, picture from \citeA{elman1990finding}}\label{fig:SRN}
    \end{subfigure}
    \begin{subfigure}[b]{0.49\linewidth}
        \centering
        \includegraphics[width=0.9\textwidth]{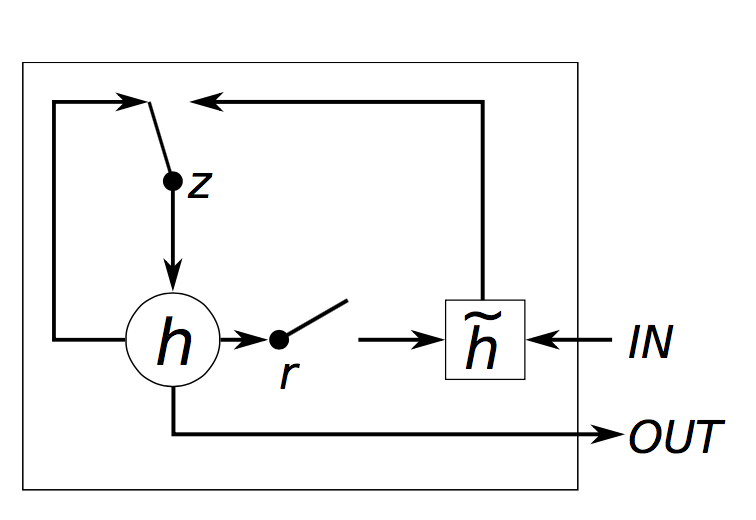}
        \caption{GRU, picture from \citeA{chung2014empirical}}\label{fig:GRU}
    \end{subfigure}
    \caption{Simple and gated recurrent neural network}\label{fig:rnns}
\end{figure}

An SRN (see Figure~\ref{fig:SRN}) consists of a single hidden layer $\vec{h}$ with a nonlinear activation function and a feedback connection that introduces a simple form of memory:

\begin{equation}
    \vec{h}_t = \tanh(\mat{W}\vec{x}_t + \mat{U}\vec{h_{t-1}} + \vec{b}),
\end{equation}

\noindent where the weight matrix $\mat{W}\in \mathbb{R}^{h\times d}$ connects the input $\vec{x}\in \mathbb{R}^d$ to the hidden layer $\vec{h}\in\mathbb{R}^h$ and the weight matrix $\mat{U}\in\mathbb{R}^{h\times h}$ stores the recurrent connections of the hidden layer to itself.

\subsubsection{GRU}
The GRU (see Figure~\ref{fig:GRU}) is an extended version of this model, in which \textit{gates} are used to modulate the information flow. 
Two gate values are computed from the previous hidden layer activation and the current input. 
The \emph{reset} gate $\vec{r}\in\mathbb{R}^h$ is used to compute a candidate hidden layer state for the next time step $\tilde{\vec{h}}_t$:

\begin{equation}
    \tilde{\vec{h}}_t = \tanh(\mat{W}\vec{x}_t + \mat{U}_h(\vec{r}_t\odot\vec{h}_{t-1}) + \vec{b}_h),
\end{equation}

\noindent where $\odot$ denotes the element wise product.
The gate value $\vec{r}$ thus determines to what extent the hidden layer activation in the previous time step $\vec{h_{t-1}}$ should be taken into account before the squashing function tanh is applied, and is computed considering the previous activation as well as the current input, using weight matrices $\mat{W_r}\in \mathbb{R}^{h\times d}$, $\mat{U_r}\in\mathbb{R}^{h\times h}$ and bias $\vec{b_r}\in\mathbb{R}^h$:

\begin{equation}
    \vec{r}_t = \sigma(\mat{W}_r\vec{x}_t + \mat{U}_r\vec{h}_{t-1} + \vec{b}_r),
\end{equation}

where $\sigma$ denotes the logistic sigmoid function.
The update gate $\vec{z}\in\mathbb{R}^h$ then determines the proportion of a hidden unit's previous activation $\vec{h}_{t-1}\in\mathbb{R}^h$ that is retained, and how much it is updated with $\tilde{\vec{h}}_t$, without using a non-linear squashing function:

\begin{equation}
        \vec{h}_t = \vec{z}_t\odot\vec{h}_{t-1} + (1-\vec{z}_t)\odot\tilde{\vec{h}}_t,
\end{equation}

\noindent where $\vec{z}$, associated with its own weight matrices $\mat{W_z}$, $\mat{U_z}$ and bias $\vec{b_z}$, is computed analogously to the other gate:
\begin{equation}
    \vec{z}_t = \sigma(\mat{W}_z\vec{x}_t + \mat{U}_z\vec{h}_{t-1} + \vec{b}_z)
\end{equation}

All weight matrices $\mat{W}\in \mathbb{R}^{h\times d}$ thus connect the input $\vec{x}\in\mathbb{R}^d$ to a gate or the hidden layer, respectively, and the weight matrices $\mat{U}\in \mathbb{R}^{h\times h}$ connect the gates to the hidden layer and the hidden layer to itself.

%%%%%%%%%%%%%%%%%%%%%%%%%%%%%%%%%%%%%%%%%%%%%%%%%%%%%%%%%%%%%%%%%%%%%%%%%%%%%%%
% Data and Results

\section{Training and Performance}\label{sec:training}
We run a series of experiments training several models on a randomly sampled set of expressions from $\mathbf{L1}$, $\mathbf{L2}$, $\mathbf{L4}$, $\mathbf{L5}$ and $\mathbf{L7}$ (3000 expressions from each subset). 
The composition of lengths of the training set remains the same across all experiments, but which exact expressions of these five lengths are chosen depends on the random seed during initialisation of the experiment.
We set the input dimensionality of the word embeddings to 2. 
The word embeddings are randomly initialised within a small range, and are updated during training.
All models are trained using backpropagation with minibatches of size 24. 

We test on a large sample of expressions from $\mathbf{L1}$, $\mathbf{L2}$, \dots, $\mathbf{L9}$.
All test sets but $\mathbf{L1}$ and $\mathbf{L2}$ contain expressions unseen during training, but testing also on expressions from $\mathbf{L3}$, $\mathbf{L6}$, $\mathbf{L8}$ and $\mathbf{L9}$ allows us to test also for generalisation to both longer and shorter unseen \textit{structures}.

\subsection{TreeRNN}
The TreeRNN models are trained using a training regime inspired by \citeA{bowman2015recursive}.
In this regime, the composition function of the network is used to compute the meaning representations of two distinct sentences.
These representations are then given as input to an additional neural classifier, that predicts the relation between the concatenated sentence representations (see Figure~\ref{fig:comparisonTraining}).
In our case, this relation is the (in)equality (`$<$', `$=$' or `$>$') that holds between their solutions.
The parameters (i.e.\ the composition matrix and the word embeddings) are updated through the optimisation of the performance of this classifier, using Adagrad \cite{zeiler2012adadelta} and a cross-entropy loss.

\paragraph{Results} Out of the 8 TreeRNNs trained, 6 obtain a high classification accuracy between 0.98 and 0.99 on the test data (sentences up to length 9), the other 2 models perform at chance level.
We will return to a more in depth analysis of the results later.

\subsection{RNN}
The sequential models are not provided with explicit structural information, but have access to the syntactic structure implicitly via the brackets, which are presented to the network as words.
As a consequence, the sentences processed by the sequential models are substantially longer than the ones presented to the recursive network (e.g. 33 versus 17, respectively, for an $\mathbf{L9}$ sentence).
All reported models are trained on an error signal backpropagated from a simple linear perceptron predicting the real-valued solution of the expressions, using Adam \cite{kingma2014adam} as optimiser (learning rate=0.001, $\beta_1$=0.9, $\beta_2$=0.999, $\epsilon$=1e-08, decay=0.0) and mean squared error as loss function. 
The hidden layer dimensionality of all recurrent models is set to $15$.
For the implementation we use the Python library Keras \cite{chollet2015} with Theano \cite{theano2016short} as backend.\footnote{The Python package containing the source code of this project can be found at \url{https://github.com/dieuwkehupkes/processing_arithmetics}.}

\begin{figure}[h]
    \centering
    \begin{tikzpicture}
  \centering
  \begin{axis}[
        height=8cm, width=\linewidth,
        symbolic x coords={L1, L2, L3, L4, L5, L6, L7, L8, L9},
        % ybar,
        % bar width=4pt,
        ymin=-0.2, ymax=100,
        axis x line*=bottom,
        axis y line*=left,
        ymajorgrids,
        grid style={loosely dashed, white!55!black!45},
        y axis line style={opacity=0},
        x axis line style={opacity=0},
        tickwidth=0pt,
        xtick=data,
        xticklabels={\textcolor{gray}{L1}, \textcolor{gray}{L2}, \textbf{L3}, \textcolor{gray}{L4}, \textcolor{gray}{L5}, \textbf{L6}, \textcolor{gray}{L7}, \textbf{L8}, \textbf{L9}},
        tick label style={font=\large},
        ylabel={mean squared error},
        legend style={at={(0.05, 0.95)}, anchor=north west}
    ]

 % GRU averages 
    \addplot[color=green!70!black, thick, mark=o, mark options={scale=1.5}, error bars/.cd, y dir=both, y explicit, error bar style={color=green!60!black}] coordinates{
     (L1,  0.02) +- (0, 0.0)
     (L2,  0.08) +- (0, 0.0)
     (L3,  0.24) +- (0, 0.1)
     (L4,  0.52) +- (0, 0.2)
     (L5,  1.61) +- (0, 0.5)
     (L6,  4.05) +- (0, 1.2)
     (L7,  9.27) +- (0, 2.3)
     (L8,  16.9) +- (0, 3.5)
     (L9,  28.5) +- (0, 5.1)
 };
 \addlegendentry{GRU average}

% GRU best 
 \addplot[green!70!black, mark=*, mark options= {solid, scale=1.5}, ultra thick] coordinates{
    (L1, 0.00) 
    (L2, 0.03)
    (L3, 0.11)
    (L4, 0.20)
    (L5, 0.57)
    (L6, 1.54)
    (L7, 3.74)
    (L8, 7.37)
    (L9, 13.57)
};
\addlegendentry{GRU best}

% SRN averages  % 3 non generalising models removed
\addplot+[densely dotted, thick, color=blue, mark=square, mark options = {solid, scale=1.5}, error bars/.cd, y dir=both, y explicit, error bar style={color=blue}] coordinates{
    (L1, 0.21) +- (0, 0.1 ) 
    (L2, 1.52) +- (0, 0.4 ) 
    (L3, 13.3) +- (0, 5.0 ) 
    (L4, 20.0) +- (0, 6.7 ) 
    (L5, 37.5) +- (0, 10.7) 
    (L6, 59.7) +- (0, 15.0) 
    (L7, 86.6) +- (0, 18.4) 
    (L8, 116.) +- (0, 20.5) 
    (L9, 157.) +- (0, 23.3) 
};
\addlegendentry{SRN average}

% SRN best % TODO replace with SRN results
\addplot+[color=black!30!blue!70, mark=square*, mark options = {solid, scale=1.5}, densely dotted, ultra thick] coordinates{
    (L1, 0.08)
    (L2, 0.34)
    (L3, 0.91)
    (L4, 2.02)
    (L5, 3.72)
    (L6, 7.18)
    (L7, 14.5)
    (L8, 24.8)
    (L9, 40.6)
};
\addlegendentry{SRN best}

\end{axis}

\end{tikzpicture}
    \caption{Average and best mean squared error for 20 GRU and 17 SRN models. Three other trained SRN models that did not learn to capture any structural knowledge were excluded from the plot. 
    Error bars indicate standard error.
    Sentences of lengths that were not included in the training set are bold-faced.}\label{fig:rnn_results}
\end{figure}
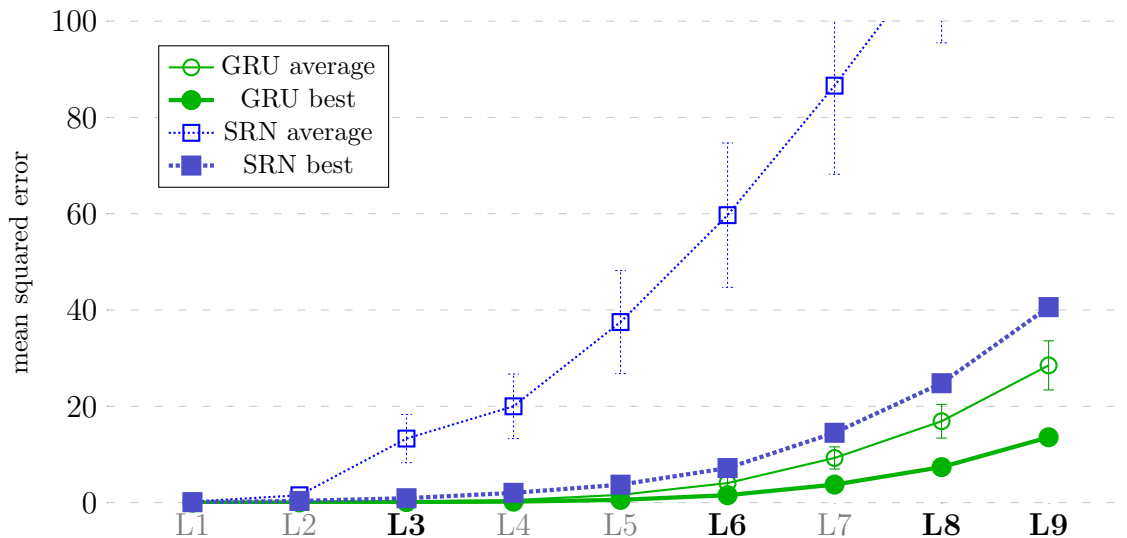

\paragraph{Results} We train 20 SRN and 20 GRU model instances, that differ with respect to both weight initialisation and exact training instances.
Of the 20 trained SRN models, three did not learn to capture any structural knowledge, reflected by a high error for short (but unseen) sentences with three numerals ($\mathbf{L3}$).
It is unclear to what extent the remaining 17 SRN models learned solutions incorporating the syntactic structure of sentences.
Most GRU models, on the other hand, show a convincing ability to generalise, with a mean squared error that slowly increases with the length of the sentence. 
A summary of the results is plotted in Figure \ref{fig:rnn_results}.

\subsection{Discussion}
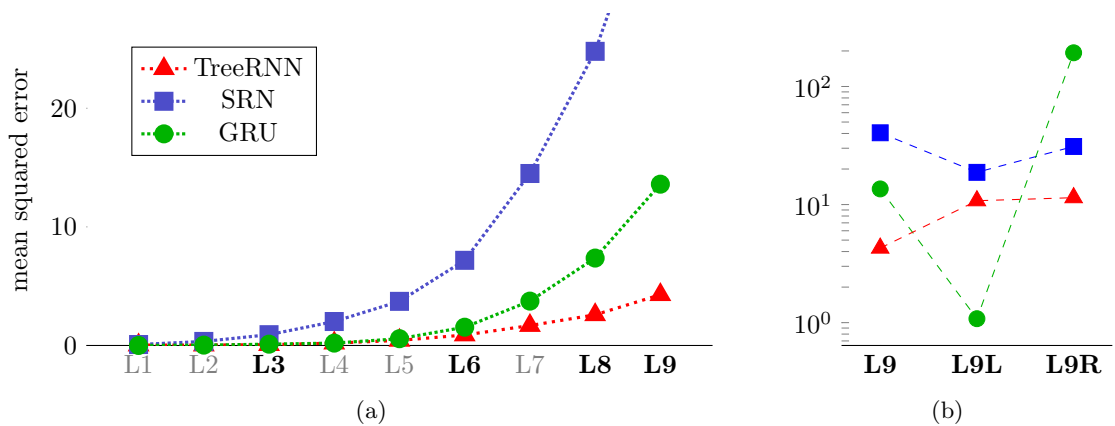
\begin{figure}[h]
\centering
\begin{subfigure}[t]{0.65\linewidth}
%% mean squared error all models

\begin{tikzpicture}
  \centering
  \begin{axis}[
        height=6cm, width=\linewidth,
        symbolic x coords={L1,L2,L3,L4,L5,L6,L7,L8,L9},%,L9L,L9R},
        ymin=0, ymax=28,
        axis x line*=bottom,
        axis y line*=left,
        y axis line style={opacity=0},
        tickwidth=0pt,
        xtick=data,
        xticklabels={\textcolor{gray}{L1}, \textcolor{gray}{L2}, \textbf{L3}, \textcolor{gray}{L4}, \textcolor{gray}{L5}, \textbf{L6}, \textcolor{gray}{L7}, \textbf{L8}, \textbf{L9}},
        ylabel={mean squared error},
        legend style={at={(0.22,0.9)},
        anchor=north},
    ]

    \addplot[color=red, dotted, mark=triangle*, mark options={solid, scale=1.8}, very thick] coordinates{
%% TRNN d2 comparison 2 hidden 4
%% model 304-205
%%  mean squared error:
(L1,	0.032995844)
(L2,	0.031150039)
(L3,	0.069888369)
(L4,	0.197978194)
(L5,	0.429466967)
(L6,	0.882411527)
(L7,	1.678663808)
(L8,	2.586690488)
(L9,	4.279257817)
};
%\addplot[color=teal,only marks,mark=x,mark options={solid},label=TreeRNN] coordinates{
%(L9L,	10.7766808)
%(L9R,	11.43800754)
%};
%% mean squared prediction error:
%L1	0
%L2	0
%L3	0.05
%L4	0.226333333
%L5	0.4934
%L6	0.9593
%L7	1.7608
%L8	2.663266667
%L9	4.360933333
%L9L	10.84986667
%L9R	11.53326667

    % mean squared error SimpleRNN seed 0
    \addplot[color=black!30!blue!70, mark=square*, dotted, mark options = {scale=1.5, solid}, densely dotted, very thick] coordinates{
        (L1, 0.08)
        (L2, 0.34)
        (L3, 0.91)
        (L4, 2.02)
        (L5, 3.72)
        (L6, 7.18)
        (L7, 14.5)
        (L8, 24.8)
        (L9, 40.6)
    };
    \addlegendentry{SRN}
%\addplot[color=blue,only marks,mark=*,label=SRN15] coordinates{
%    (L9R, 37.6527429637)
%    (L9L, 9.60665151672)
%};

    % mean squared error GRU 
    \addplot[green!70!black, mark=*, mark options= {scale=1.5, solid}, very thick, densely dotted] coordinates{
        (L1, 0.00)
        (L2, 0.03)
        (L3, 0.11)
        (L4, 0.20)
        (L5, 0.57)
        (L6, 1.54)
        (L7, 3.74)
        (L8, 7.37)
        (L9, 13.6)
};
\addlegendentry{GRU}

%    \addplot[scatter, only marks, point meta=explicit symbolic,
%        scatter/classes={
%            LBSRN={mark=triangle,blue},
%            LBGRU={mark=triangle,red},
%            LBTree={mark=triangle,green},
%            RBSRN={mark=square,blue},
%            RBGRU={mark=square,red},
%            RBTree={mark=square,green}}
%        ]
%    table[meta=label] {
%        x    y      label
%        L9  9.61    LBSRN 
%        L9  0.90    LBGRU
%        L9  10.78   LBTree
%        L9  37.65   RBSRN
%        L9  158.13  RBGRU
%        L9  11.4    RBTree
%    };

%     \addplot+[nodes near coords,only marks,
%         point meta=explicit
%     table[meta=label] {
%         x y label
%         0.1 0.15 a
%         0.45 0.27 c
%         0.02 0.17 a
%         0.06 0.1 a
%         0.9 0.5 b
%         0.5 0.3 c
%         0.85 0.52 b
%         0.12 0.05 a
%         0.73 0.45 b
%         0.53 0.25 c
%         0.76 0.5 b
%         0.55 

\legend{TreeRNN, SRN, GRU}
      \end{axis}
  \end{tikzpicture} 
\caption{}
\label{f:resultsAll}
\end{subfigure}
\begin{subfigure}[t]{0.34\linewidth}
    \centering
 \begin{tikzpicture}
  \centering
  \begin{axis}[
        height=5.5cm, width=\linewidth,
        symbolic x coords={L9,L9L,L9R},
        ymin=0, ymax=200,
ymode=log,
        axis x line*=bottom,
        axis y line*=left,
       y axis line style={opacity=0},
        tickwidth=0pt,
        xtick=data,
        xticklabels={\textbf{L9}, \textbf{L9L}, \textbf{L9R}},
        enlarge x limits=0.2,
    ]

% SRN
\addplot[color=blue, mark=square*, mark options = {solid, scale=1.5},  dashed] coordinates{
    (L9, 40.55)
    (L9L, 18.75)
    (L9R, 31.05)
};

% GRU
\addplot[color=green!70!black, mark=*, mark options = {solid, scale=1.5}, dashed] coordinates{
       (L9, 13.57)
       (L9L, 1.08)
       (L9R, 193)
};

% TreeRNN
\addplot[color=red, mark=triangle*, mark options = {solid, scale=1.9}, dashed] coordinates{
    (L9, 4.279257817)
    (L9L, 10.7766808)
    (L9R, 11.43800754)
};

\end{axis}
\end{tikzpicture}
\caption{}
\label{f:resultsBranching}
\end{subfigure}
\caption{(a) Mean squared error of the best performing SRN, GRU and TreeRNN models tested on sentences with a different number of numerals. (b) Mean squared error (log scale) of the best performing models tested on left- and right branching sentences containing 9 numerals ($\mathbf{L9L}$ and $\mathbf{L9R}$, respectively). Sentence lengths not included in the training set are bold-faced. }\label{fig:results_plot} 
\end{figure}

In Figure~\ref{fig:results_plot} we plot the performances of the best TreeRNN, SRN and GRU models.\footnote{To be able to compare the results of the recursive and recurrent neural networks, we trained a neural network with one hidden layer to predict the meaning of sentences from the representations generated by the best TreeRNN.}
We find that the TreeRNN generalises adequately to longer sentences, evident from the smooth progression of the performance with respect to length \cite<a criterion used by>{bowman2015treestructure}.
Although the generalisation capacity of the GRU (with $|h|=15$) is weaker, also the GRU appears to have found a solution that is sensitive to the syntactic structure. 
From these results, it is unclear to what extent the best SRN model has learned a solution incorporating the syntactic structures of sentences.

A striking difference between the GRU and TreeRNN appears when considering their performance on sentences that are fully left- or right-branching (see Figure~\ref{fig:results_plot}).
While fully right-branching sentences are much harder to process for the GRU models, the performance of TreeRNNs suffers more from tree \textit{depth}, regardless of the branching direction.
We revisit this topic later.

%%%%%%%%%%%%%%%%%%%%%%%%%%%%%%%%%%%%%%%%%%%%%%%%%%%%%%%%%%%%%%%%%%%%%%%%%%%%%%%%
% Analysis

\section{Analysis}\label{sec:analysis}

In the remainder of this paper, we present an analysis of the internal dynamics of the best performing models.
We start by considering the TreeRNN, whose extremely low dimensionality allows us to inspect and understand the information encoded in the hidden layers and weight matrices. 
Subsequently, we analyse the internal dynamics of the GRU network, and introduce a new approach for testing hypotheses about the strategies a network might engage. 

\subsection{TreeRNN}

The recursive architecture of the TreeRNN forces the model to compute the meanings of the input sentences in a purely recursive fashion.
Consequently, we do not have to discover \emph{what} the model is doing, but only \emph{how} this is implemented in the composition function.
For this analysis, we use a method we call \textit{project-sum-squash}, in which we separately consider the three operations executed by the weight matrix \cite<see also>{veldhoen2015semantic}.
First, we consider how the three input components (numeral, operator, numeral) are \emph{projected} by the composition matrix to different positions in the two dimensional state space. Next, the net input to the neurons of the parent layer is effectively the \emph{sum} of these three projections and the bias. 
Lastly, we study the effect of applying the activation function of the network, which \emph{squashes} the net input back into the range $\{-1,1\}$.

\begin{figure}[!h]\center
\begin{subfigure}[b]{0.62\linewidth}
\includegraphics[width=1.0\linewidth, trim=23mm 35mm 19.5mm 50mm, clip]{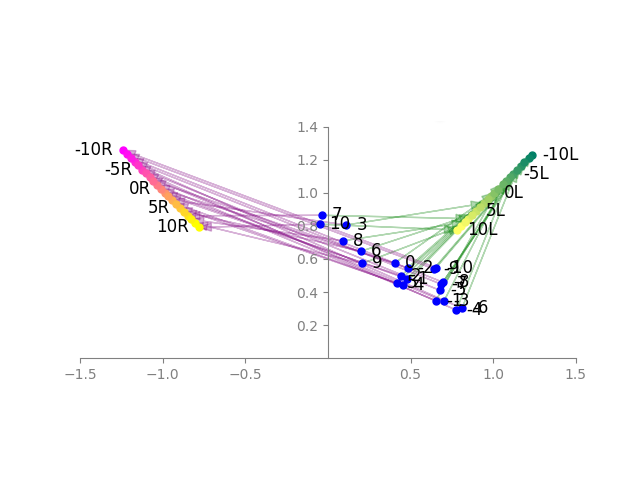}
\caption{}
\label{fig:word_embeddings}
\end{subfigure}
\begin{subfigure}[b]{0.37\linewidth}
    \centering
    \includegraphics[width=1.05\linewidth, trim=30mm 20mm 11mm 20mm, clip]{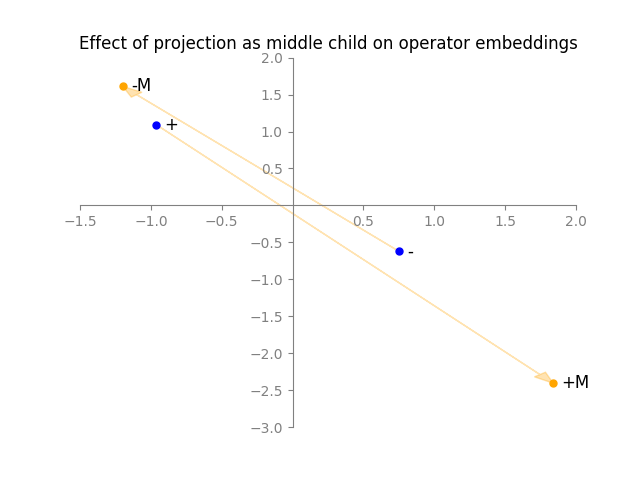}
    \caption{}\label{fig:operator_embeddings}
\end{subfigure}
\caption{Trained word embeddings and their projections. In (a) the projections of the seemingly unorganised numeral embeddings (blue) through $\mat{W}_L$ (as left child, green) and $\mat{W}_R$ (as right child, purple) reveal their organisation. In (b) we see the embeddings of the operators \texttt{plus} and \texttt{minus} and their projections after multiplication with $\mat{W}_M$ (orange).}
\end{figure}

\paragraph{Project}
As the weight matrix $\mat{W}$ can be written as a concatenation $[\mat{W}_L;\mat{W}_M;\mat{W}_R]$ -- where each $\mat{W}_i \in \mathbb{R}^{2\times 2}$ -- we can consider the projections of the three input children individually.
Figure~\ref{fig:word_embeddings} depicts the numeral embeddings and their projections, illustrating how the initially seemingly unorganised numerals embeddings are projected to organised, almost orthogonal subspaces, depending on whether they are the left or right child.
The projections of the plus and minus operator are roughly opposite to each other (see Figure~\ref{fig:operator_embeddings}).

\begin{figure}[!h]\center
\begin{subfigure}{.58\linewidth}
\includegraphics[width=0.9\linewidth, trim=55mm 28mm 17mm 29mm, clip]{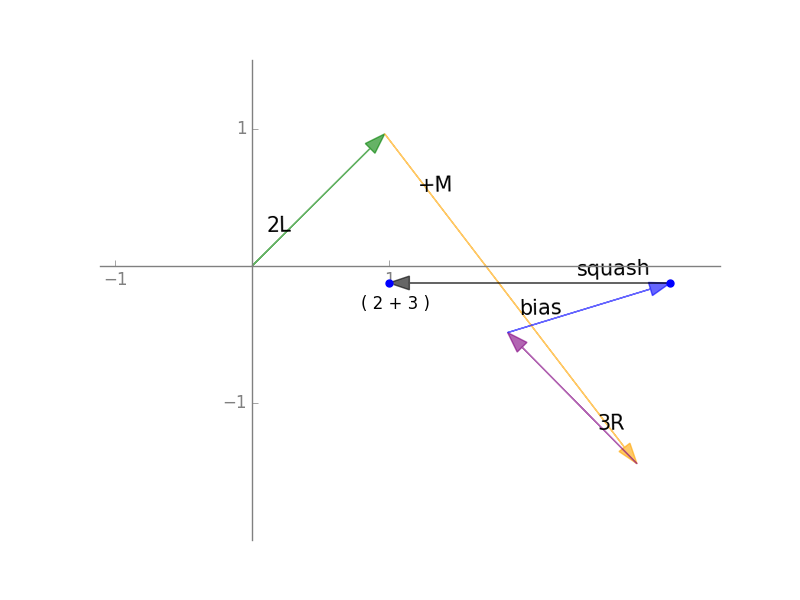}
\caption{{\tt ( 2 + 3 )}}
\label{f:treeExampleA}
\end{subfigure}
\begin{subfigure}{.41\linewidth}
\includegraphics[width=0.9\linewidth, trim=70mm 39mm 50mm 13mm, clip]{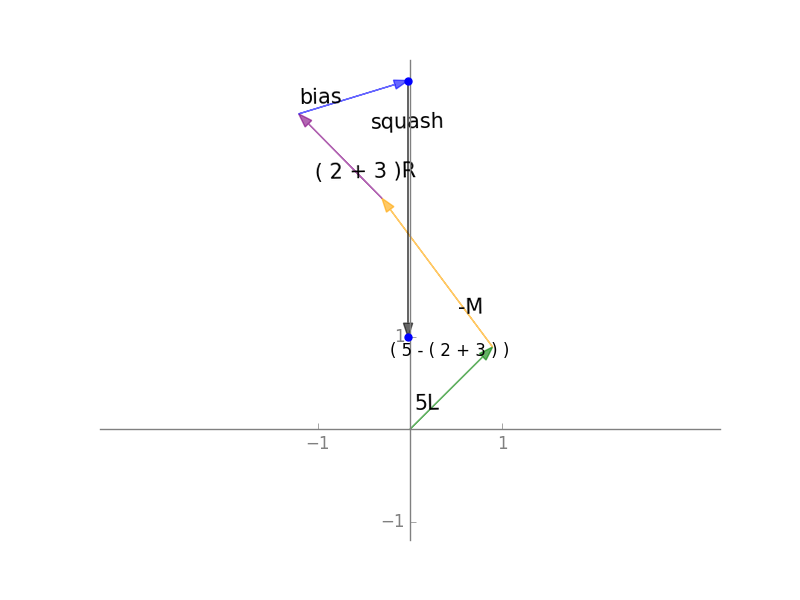}
\caption{{\tt ( 5 - `( 2 + 3 )' )}}
\label{f:treeExampleB}
\end{subfigure}
\caption{Summation of projections -- left (green), middle (yellow) and right (purple) child -- and the bias term (blue). Squash ($\tanh$ activation, grey) shifts the result back to the range $(-1, 1)$.}
\label{fig:tree_example}
\end{figure}

\paragraph{Sum} After the representations are projected, they are summed and the bias term is added, as depicted by the coloured arrows in Figure~\ref{fig:tree_example}.
Due to the approximately orthogonal and relatively small subspaces the projections of the projected left and right representations inhabit, the result is a position close to one of the axes.
The operator determines whether this is the x-axis (for \texttt{plus}) or y-axis (for \texttt{minus}).

\paragraph{Squash} In the last step, both coordinates of the representation are squashed into the range $(-1, 1)$ by the activation function {\tt tanh}.
After applying the composition matrix, the sum of the projected representations is already within this range for one of the dimensions and well above it for the other (which depends on the operator).
The net effect of the squash function is therefore a horizontal or vertical shift.
Figure~\ref{fig:L9_representations} shows the representations of 2000 $\mathbf{L9}$ sentences, showing that this effect remains constant when the length of expressions grows.

\subsubsection{Recursion} 
\begin{figure}[!h]\center
\begin{subfigure}{.49\linewidth}
\includegraphics[width=1.1\linewidth, trim=2cm 2cm 25mm 9mm, clip]{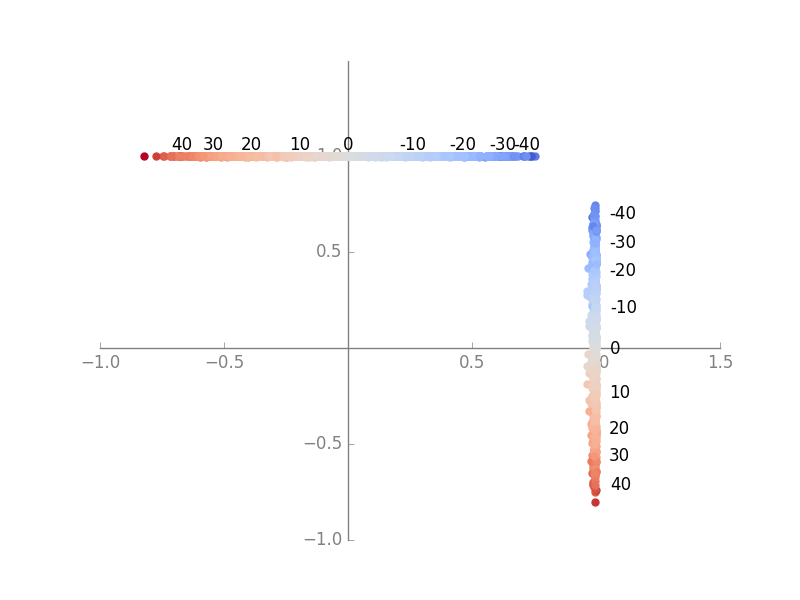}
\caption{}
\label{fig:L9_representations}
\end{subfigure}
\begin{subfigure}{.49\linewidth}
\includegraphics[width=\linewidth, trim=2cm 2cm 2cm 2cm, clip]{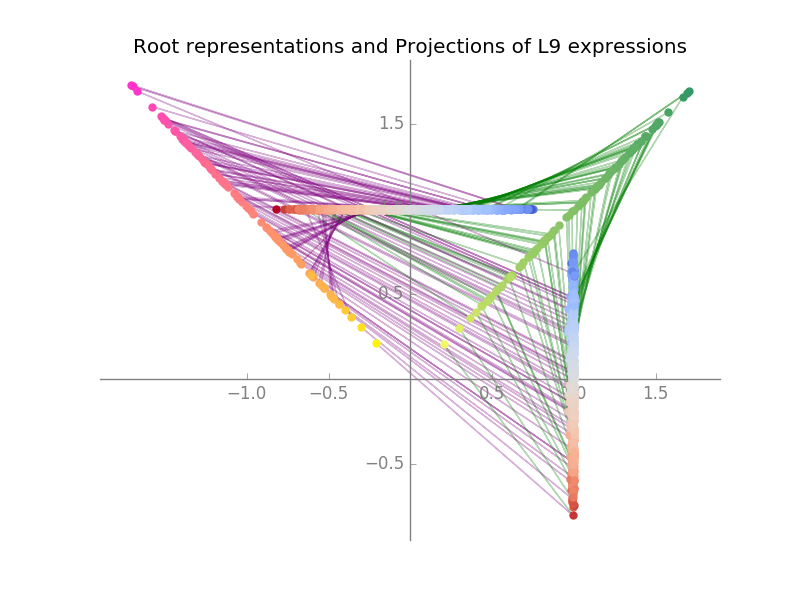}
\caption{}
\label{fig:L9_projections}
\end{subfigure}
\caption{(a) Representations of 2000 $\mathbf{L9}$ expressions whose root operation was addition ($x\approx 1$, vertical line) or subtraction ($y\approx 1$, horizontal line). 
(b) The projections of the same 2000 expressions as left and $\mat{W}_L$ (green) right $\mat{W}_R$ (purple) child when they serve as input to a new computation.
}
\label{f:treeRecursion}
\end{figure}

In the \emph{project} and \emph{squash} steps it becomes clear that the initial word embeddings and the vectors representing larger phrases do not inhabit the same subspace.
Where the word embeddings exist on a seemingly unorganised diagonal line (see Figure~\ref{fig:word_embeddings}), the larger phrase representations exist on horizontal ($y=1$) and vertical lines ($x=1)$.
However, Figure~\ref{fig:L9_projections} illustrates that the left and right projections of both phrase and word representations do get projected onto the same orthogonal subspaces.
This means the composition function can in theory be applied recursively for an indefinite number of steps.

\paragraph{} In summary, the project-sum-squash analysis nicely illustrates the process by which the TreeRNN recursively computes the compositional meaning of sentences in the arithmetic language.

\subsection{RNN Models}

Although insightful for the extremely low dimensional recursive networks, the visualisation based project-sum-squash method becomes increasingly unpractical when the dimensionality of the  networks increases.
Additionally, the sequential structure of the RNN models and the complex interaction between gates and hidden states poses difficulties for the analysis of their internal dynamics.
While the TreeRNN models process their input recursively by definition, the sequential models are forced to follow an incremental protocol to process the recursively structured sentences, which makes it much harder to discover  which operation is carried out at which point in time.  In other words, for the TreeRNN models we had to figure out \emph{how} models are implementing the operations to compute the meaning of an expression, for the RNN models we also need to determine \emph{what} these operations are.  A sequential equivalent of the project-sum-squash method is thus difficult to realise, and for the analysis of the dynamics of our RNN models we have to resort to different methods.  

In this section we first briefly discuss and test some previously proposed methods for visualising the internal state dynamics of neural networks, described in papers with an objective similar to ours \cite<e.g.>[all present detailed analyses of gated RNNs performing tasks involving compositionality]{karpathy2015visualizing,li2016visualizing,tang2017memory}.
Subsequently, we discuss the shortcomings of these methods and propose an approach we call `diagnostic classification' to test hypotheses on which strategy a trained neural network is effectively following. 
At the end of this section, we apply this method to interpret the inner workings of the best performing GRU model trained on the arithmetic language. 

\subsubsection{Visualising Neural Networks}

Understanding the internal dynamics of recurrent neural networks is a topic that is receiving more and more attention. Most papers we are aware of concern visualisation of properties of internal states of RNNs.
For example, \citeA{karpathy2015visualizing} and \citeA{li2016visualizing} study cell and gate activations under different conditions; \citeA{tang2017memory} plot distributions of cell activations and temporal traces of t-SNE cell vectors of a speech recognition RNN; and \citeA{strobelt2016visual} present an elaborate tool\footnote{LSTMVis, \url{http://lstm.seas.harvard.edu/}, is a visualisation tool that can be applied to gated RNNs, including the GRU model.} that facilitates visual analysis of gated RNNs.
We briefly discuss some of their methods and results and employ them to analyse our own models.

\begin{figure}[ht]
    \setlength{\tabcolsep}{-1pt}
    \begin{tabular}{lc}
        \specialcell{\includegraphics[width=0.920\linewidth, trim=0mm 75mm 7mm 62mm, clip]{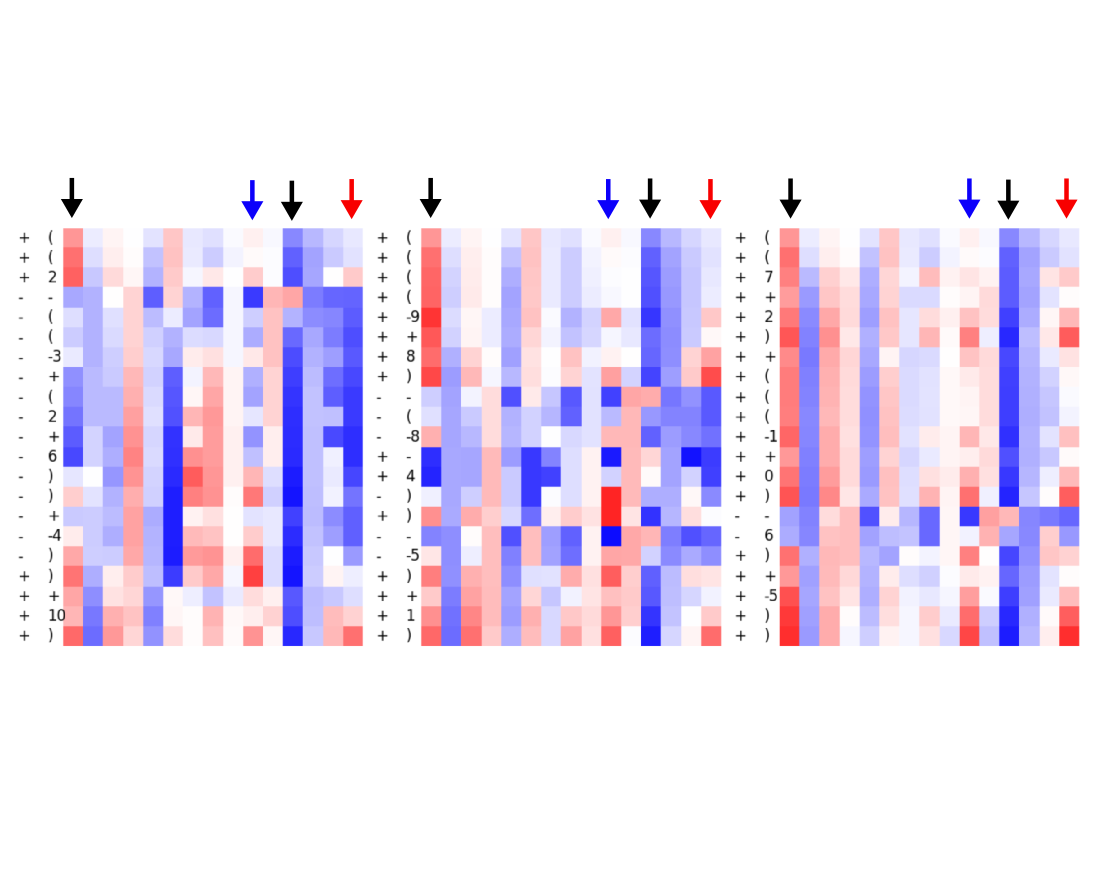}} & \specialcell{\includegraphics[scale=0.44, trim=158mm 12mm 0cm 0cm, clip]{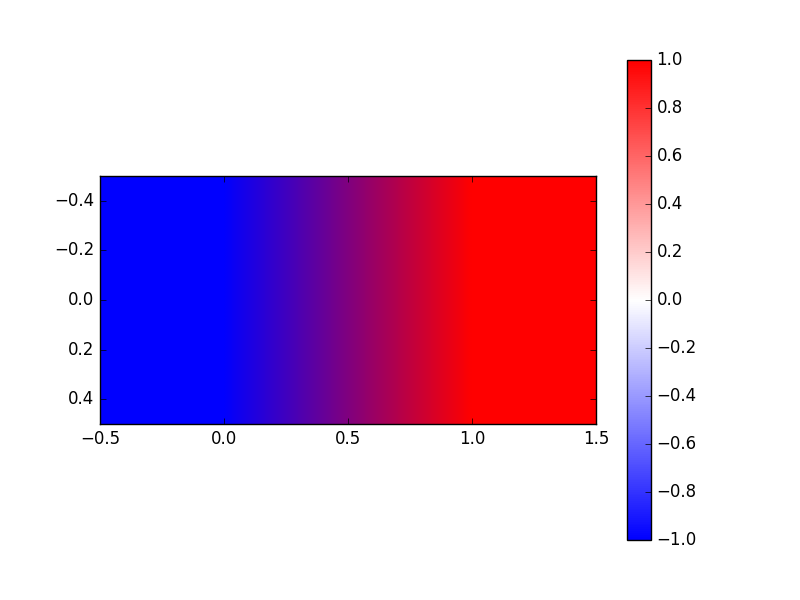}}\\
        \hspace{7mm} \includegraphics[width=0.275\linewidth, trim=25mm 70mm 7mm 71mm, clip]{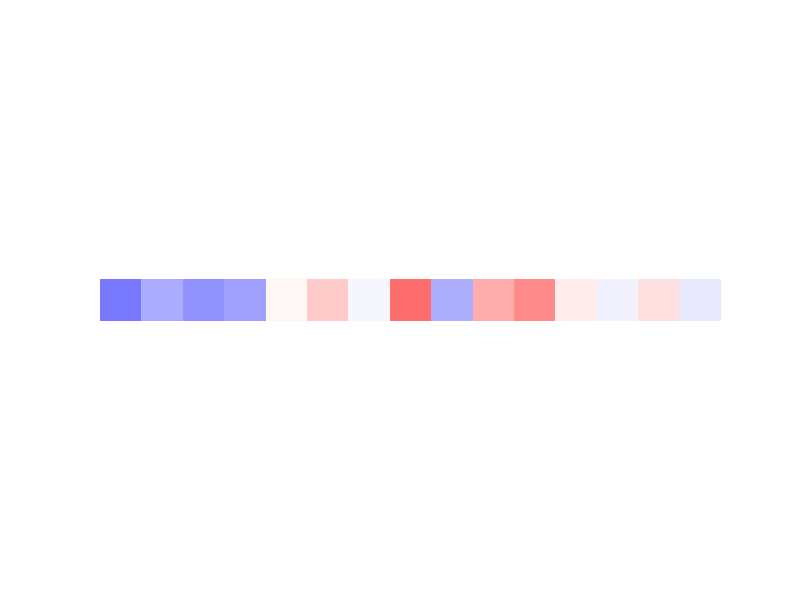} 
        \hspace{2mm} \includegraphics[width=0.275\linewidth, trim=25mm 70mm 7mm 71mm, clip]{figures/output_classifier.png} 
        \hspace{2mm} \includegraphics[width=0.275\linewidth, trim=25mm 70mm 7mm 71mm, clip]{figures/output_classifier.png} 
    \end{tabular}
    \caption{Hidden layer activations of a trained GRU network while processing different sequences.
        The input labels, along with the \texttt{mode} (addition/subtraction) at every point in time are printed left of the activation values.
        The activation values of cell 1 and 12 (black arrows) seem to be correlated with the \texttt{mode}, but it is is not easy to determine whether this value is in fact computed by the network. The 10th cell could be keeping track of the scope of a minus.}
\label{fig:hl_activations}
\end{figure}

\paragraph{Individual cell dynamics}
A potential strategy to gain a better understanding of the internal dynamics of a recurrent neural network is to consider the interpretability of individual cells.
By plotting individual cell activations, \citeA{karpathy2015visualizing} discovered several interpretable cells in a character-based neural language model, including cells that keep track of the scope of quotes and cells that represent the length of the sentence. We here apply this approach to the best performing GRU model from our experiments.
In Figure~\ref{fig:hl_activations} we plot the hidden layer activations while the network is processing three different input sequences.
Next to the activations, we put both the input sequence (starting from the top) as well as the \texttt{mode} of operation of the cumulative strategy in Figure~\ref{fig:pseudo} (i.e.\ \texttt{+} or \texttt{-}) at each point in time.
Under the hidden layer activations, we also plot the weights of the output layer reading out the meaning of the sentence, using the same colour scale as for the hidden layers (plotted at the right of the figure).

From this picture, several interesting observations can be made.
Both the first cell (the first column from the left in all three graphs, black arrow) and the twelfth cell (the fourth column from the right, black arrow) show a sharp change in activation whenever the mode changes from \texttt{+} to \texttt{-}.
The last layer of the network (bottom of the figure) indicates that the leftmost cell is negatively influencing the prediction of the solution of the expression, while the twelfth cell is hardly involved.
The very last cell (red arrow) seems to respond to a minus in the input, but appears to have also other functions.
The tenth cell (blue arrow) could potentially be representing the scope of a minus.

In summary, studying hidden layer activations is an interesting puzzle and can -- especially for relatively low dimensional network such as ours -- give pointers to which aspects should be studied in more depth.
However, it is hard to draw definite (and quantitative) conclusions, and the usefulness of the method decreases with higher dimensionality of the networks.

\paragraph{Gate activation statistics}
In the same paper, \citeA{karpathy2015visualizing} study gate activations, focusing in particular on the fraction of time that gates spend being left- or right-saturated (activations less 0.1 than or more than 0.9, respectively).
Such results are not easy to interpret, but might indicate what roles specific cells play in the information processing a network performs: Are they mostly acting as a memory, in a feed forward fashion, or as standard recurrent cells without any additional form of memory? 
For example, a cell with a right-saturated update gate remembers its previous activation, whereas a cell with a left-saturated update gate and a left-saturated reset gate ignores all previous activations and bases its value only on the current input.

\begin{figure}[h]
    \begin{subfigure}{0.49\linewidth}
    \includegraphics[width=1.1\linewidth]{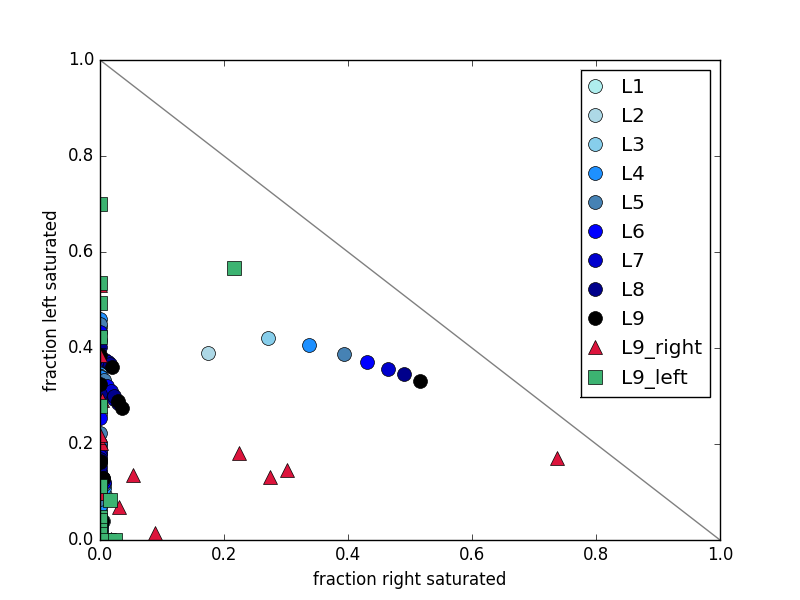}
    \caption{Reset gate $\vec{r}$ (Equation 4)}
\end{subfigure}
    \begin{subfigure}{0.49\linewidth}
    \includegraphics[width=1.1\linewidth]{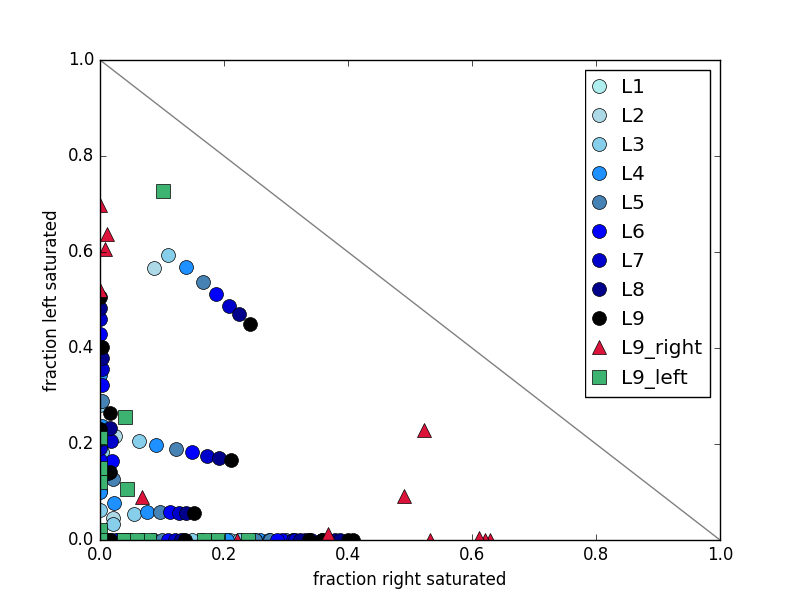}
    \caption{Update gate $\vec{z}$  (Equation 6)}
\end{subfigure}
\caption{Gate activation statistics of the update and reset gate of a GRU model while processing sentences from different lengths.
    Following \citeA{karpathy2015visualizing}, each circle represents a gate in the GRU\@.
Its position is determined by the fraction of the time it is left- (activation value less than 0.1) or right-saturated (activation value more than 0.9).}\label{fig:gate_statistics}
\end{figure}

In Figure~\ref{fig:gate_statistics} we plot the left- and right-saturation statistics for a GRU network for different lengths of expressions.
As the dimensionality of our network is considerably lower than that of the network considered by \citeauthor{karpathy2015visualizing}, we can easily visualise the gate saturation values for different sentence lengths in the same plot.
We observe that most update gates are either on the x- or y-axis. 
Some cells (at the right of the picture) act as a memory a substantial amount of the time.
A few cells show an interesting context dependency, spending an increasing fraction of the time being right-saturated.
The reset gate saturation values show that several cells spend a considerable amount of time in `feedforward-mode' (indicated by a high fraction of left-saturation).
For one cell, the activation appears to be dependent on the length of the expression. 

\subsubsection{Diagnostic Classification}

Although visualisation-based methods give intriguing clues, they do not result in a comprehensive view of the internal dynamics of a network.
The potential conclusions that can be drawn from visual inspection concern only small parts of the network's overall behaviour and are often qualitative rather than quantitative, as manually analysing the behaviour of cells over a large number of examples is infeasible.
Additionally, the described visualisation methods are restricted to finding functions or features that are encoded by one cell, while being insensitive to operations distributed over multiple cells, or cells that encode multiple features at the same time. Disentangling the behaviour of networks through visual inspection of activations is searching for a needle in a haystack.

\begin{figure}[]
\includegraphics[width=\linewidth, trim=6cm 7cm 6cm 6cm, clip]{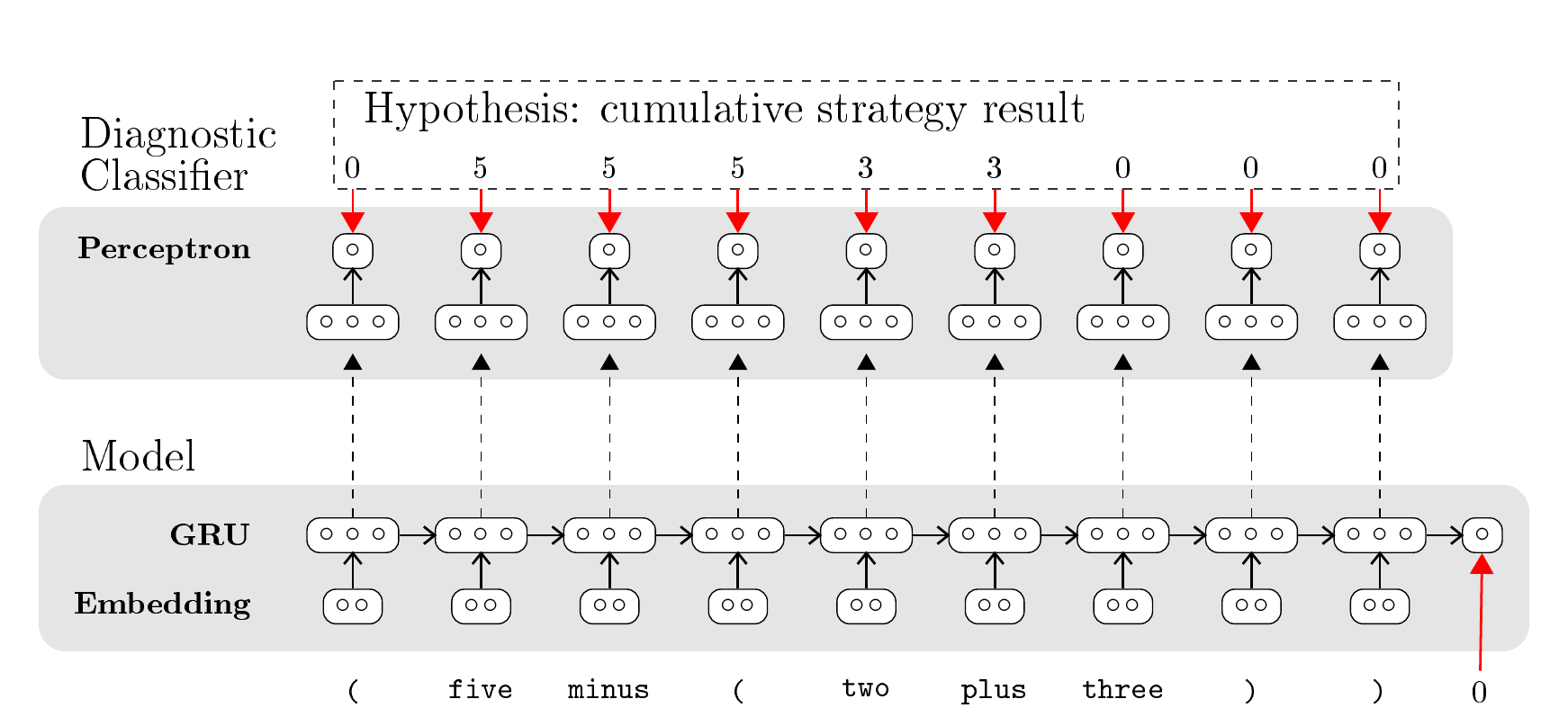}
\caption{Testing a single hypothesis with a diagnostic classifier.}\label{fig:seq2seq}
\end{figure}

We here develop an alternative approach that we call `diagnostic classification'.
Our approach is based on the idea that if a sequential model is computing certain information, or merely keeping track of it, it should be possible to extract this information from its internal state space.
To test whether a network is computing or representing a certain variable or feature, we determine the sequence of values that this variable or feature should take at each step while processing a sentence.
We then train an additional classifier -- a \emph{diagnostic classifier} -- to predict this sequence of variable values (representing our hypothesis) from the sequence of hidden representations a trained network goes through while processing the input sentence.
If the sequence of values can be predicted with a high accuracy by the diagnostic classifier, this indicates that the hypothesised information is indeed computed by the network. Conversely, low accuracy suggests this information is not represented in the hidden state. A sketch of the diagnostic classification method applied to test a hypothesis is depicted in Figure~\ref{fig:seq2seq} (which we later discuss in more detail).

Diagnostic classification is a generic method that addresses most of the shortcomings we listed for existing methods. It can be used to quantitatively test hypotheses about neural networks that range from very simple to fully fledged (symbolic) strategy descriptions.
This approach, which bears similarities with analysis methods presented by \citeA{adi2016fine} and \citeA{gelderloos2016phonemes}, can be used to quantitatively test for the existence of feature detectors such as the inside-quote detectors found by \citeA{karpathy2015visualizing}, but also can be extended to test whether a network is computing the type of information needed for the symbolic strategies defined in Figure~\ref{fig:pseudo}. 

\paragraph{Feature testing} Diagnostic classifiers can test whether a network has an internal representation of certain features of its input. 
For instance, to evaluate whether a network is keeping track of the length of a sentence \cite<another example taken from>{karpathy2015visualizing}, we can simply train a diagnostic classifier to predict the value of this variable at each point in time from the hidden state of the network while processing a corpus of sentences.
The accuracy of this diagnostic classifier will be high not only if there is a single cell acting as a length counter, but also when multiple cells are together encoding this information.
Furthermore, the accuracy with which this classifier can predict the sentence length from the sequence of hidden states gives a quantitative measure of how well this information is kept track of. 
Analysing the accuracy of the classifier in more detail (i.e.\ inspecting at which points it fails to correctly predict the sentence length) or looking at its weights can provide more insight in what the network is doing.

\paragraph{Symbolic strategies}
In a similar fashion, diagnostic classifiers can be used to probe the strategy networks could be implementing on an algorithmic level \cite{mart1982computational}, provided such strategies can be translated into sequences of targets for each time step.
The cumulative and recursive strategy we defined in Figure~\ref{cumstrat} and Figure~\ref{recstrat}, respectively, result in very different predictions about the intermediate results stored (and computed) during processing a sequence.
For instance, after seeing the word \texttt{three} in the sequence \texttt{( five minus ( ( two mines three ) plus seven ) )}, the recursive strategy should have a representation of the value within the current brackets (which is $\text{-}1$), whereas the cumulative strategy should maintain a representation of the value of the expression up to that point ($6$).
Figure~\ref{fig:seq2seq} shows the setup for training a diagnostic classifier to predict the intermediate results for the cumulative strategy.

\subsubsection{Applying Diagnostic Classifiers}

To test whether our trained networks are following either the cumulative or recursive strategy, we train diagnostic classifiers to predict the sequences of intermediate results of both these strategies, as well as the variable \texttt{mode} used by the cumulative strategy to determine whether the next number should be added or subtracted.
As the diagnostic model should merely read out whether certain information is present in the hidden representations rather than perform complex computations itself, we use a simple linear model as diagnostic classifier. 

\begin{figure}[hbt]
    \centering
    \begin{subfigure}[t]{0.32\linewidth}
        \begin{tikzpicture}
  \centering
  \begin{axis}[
        height=5cm, width=1.15\linewidth,
        ymajorgrids, tick align=inside,
        major grid style={draw=white},
        enlarge y limits={value=.1,upper},
        ymin=0, ymax=120,
        axis x line*=bottom,
        axis y line*=left,
        y axis line style={opacity=0},
        tickwidth=0pt,
        legend style={at={(0.36, 0.60)}, anchor=south},
        ylabel={mean squared error},
        symbolic x coords={
            1, 2, 3,
            4, 5, 6, 7, 8, 9, 9L, 9R},
        xtick={
            1, 2, 3,
            4, 5, 6, 7, 8, 9, 9L, 9R},
        ticklabel style = {font=\tiny},
    ]

    % Incremental strategy error GRU 8
    \addplot [dotted, color=green!60!black, mark=square*,mark options={solid}] coordinates {
        (1, 3.41)
        (2, 2.02)
        (3, 3.68)
        (4, 5.44)
        (5, 7.23)
        (6, 8.96)
        (7, 11.25)
        (8, 13.55)
        (9, 16.20)
        };

	% recursive strategy GRU8
    \addplot [dotted, color=blue, mark=diamond*, mark options={solid, scale=1.4}] coordinates {

        (1, 12.41)
        (2, 11.99)
        (3, 20.28)
        (4, 28.95)
        (5, 37.57)
        (6, 44.70)
        (7, 51.89)
        (8, 57.97)
        (9, 66.25)
        };

    \legend{cumulative, recursive}
	
    % incremental
    \addplot [color=green!60!black, mark=square*] coordinates {
        (9L, 9.11)
    };

    \addplot [color=green!60!black, mark=square*] coordinates {
        (9R, 109.36)

    };
    
    % recursive
    \addplot [color=blue, mark=diamond*, mark options={scale=1.4}] coordinates {
        (9L, 49.50)
    };

    \addplot [color=blue, mark=diamond*, mark options={scale=1.4}] coordinates {
        (9R, 115.80)
    };

  \end{axis}
  \end{tikzpicture}
        \caption{prediction of \texttt{result}}\label{fig:results_a}
    \end{subfigure}
       \begin{subfigure}[t]{0.32\linewidth}
        \begin{tikzpicture}
  \centering
  \begin{axis}[
        height=5cm, width=1.15\linewidth,
        ymajorgrids, tick align=inside,
        major grid style={draw=white},
        enlarge y limits={value=.1,upper},
        ymin=0, ymax=1.0,
        axis x line*=bottom,
        axis y line*=left,
        y axis line style={opacity=0},
        tickwidth=0pt,
        ylabel={accuracy},
        symbolic x coords={
            1, 2, 3, 4, 5, 6,
            7, 8, 9, 9R, 9L},
        xtick={
            1, 2, 3, 4, 5, 6,
            7, 8, 9, 9R, 9L},
        ticklabel style = {font=\tiny},
    ]
        
    % Incremental strategy error subtraction
    \addplot [dotted, color=green!60!black, mark=square*,mark options={solid}] coordinates {
        (1, 1.00)
        (2, 1.00)
        (3, 1.00)
        (4, 0.99)
        (5, 0.98)
        (6, 0.96)
        (7, 0.95)
        (8, 0.94)
        (9, 0.93)
        };

    \addplot [color=green!60!black, mark=square*] coordinates {
        (9L, 1.00)
    };

    \addplot [color=green!60!black, mark=square*] coordinates {
        (9R, 0.70)
    };

  \end{axis}
  \end{tikzpicture}
        \caption{prediction of \texttt{mode} cumulative}\label{fig:results_b}
    \end{subfigure}
    \begin{subfigure}[t]{0.32\linewidth}
        \begin{tikzpicture}
  \centering
  \begin{axis}[
        height=5cm, width=1.15\linewidth,
        ymajorgrids, tick align=inside,
        major grid style={draw=white},
        enlarge y limits={value=.1,upper},
        ymin=0, ymax=1,
        axis x line*=bottom,
        axis y line*=left,
        y axis line style={opacity=0},
        tickwidth=0pt,
        legend style={at={(0.2, 0.38)}, anchor=south},
        ylabel={pearsonr},
        symbolic x coords={
            1, 2, 3,
            4, 5, 6, 7, 8, 9, 9L, 9R},
        xtick={
            1, 2, 3,
            4, 5, 6, 7, 8, 9, 9L, 9R},
        ticklabel style = {font=\tiny},
    ]

    % Incremental strategy error
    \addplot [dotted, color=green!60!black, mark=square*,mark options={solid}] coordinates {
        (1, 0.99)
        (2, 0.97)
        (3, 0.97)
        (4, 0.97)
        (5, 0.96)
        (6, 0.96)
        (7, 0.96)
        (8, 0.96)
        (9, 0.95)
        };

    \addplot [color=green!60!black, mark=square*] coordinates {
        (9L, 0.98)
    };

    \addplot [color=green!60!black, mark=square*] coordinates {
        (9R, 0.72)
    };

	% recursive strategy
    \addplot [dotted, color=blue, mark=diamond*,mark options={solid, scale=1.4}] coordinates {
          (1, 0.98)
          (2, 0.91)
          (3, 0.80)
          (4, 0.73)
          (5, 0.68)
          (6, 0.62)
          (7, 0.58)
          (8, 0.55)
          (9, 0.52)
        };

        \addplot [color=blue, mark=diamond*, mark options={scale=1.4}] coordinates {
        (9L, 0.94)
    };

    \addplot [color=blue, mark=diamond*, mark options={scale=1.4}] coordinates {
        (9R, 0.27)
    };

  \end{axis}
  \end{tikzpicture}
        \caption{trajectory correlations}\label{fig:results_c}
    \end{subfigure}
     \centering
     \caption{Results of diagnostic models for a GRU model on different subsets of languages.}\label{fig:seq2seq_GRU}
\end{figure}
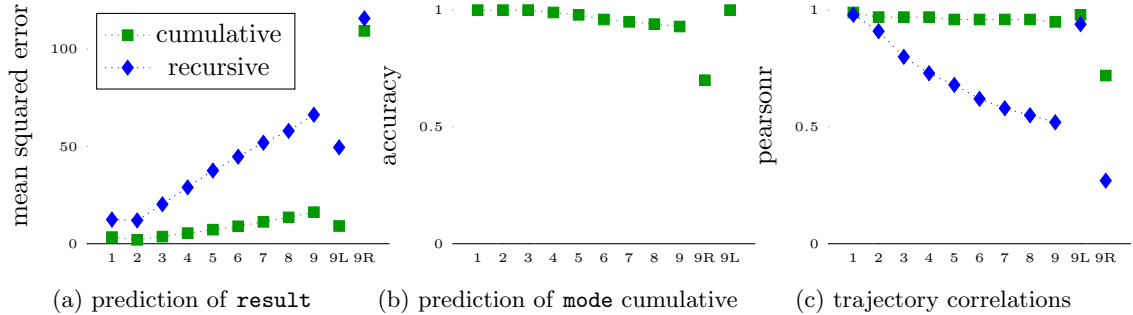

\paragraph{Strategy results}
We find that the values required for the cumulative strategy (\texttt{mode} and \texttt{result}) can be more accurately predicted than the intermediate recursive strategy values (see Figure~\ref{fig:results_a} and~\ref{fig:results_b}).
From these findings it appears unlikely that the network implements a fully recursive strategy employing a number stack of intermediate results.
For the cumulative strategy the predictions are generally accurate, even for longer sentences.
The same is true for the \texttt{mode} of subtraction of the cumulative strategy (see Figure~\ref{fig:results_b}), which can be predicted almost perfectly for sentences up until length 5 (with accuracies in the range of 0.98 -- 1.0), but is also accurately kept track of for longer sequences (an accuracy 0.93 for $\mathbf{L9}$ sentences).
However, the fit with the cumulative strategy is not perfect: the diagnostic classifiers trained to test the cumulative hypothesis perform excellently for left-branching sentences, but show low accuracy for right-branching sentences.
This is inconsistent with the symbolic description of the cumulative strategy, where the stack is crucial for left-branching sentences, but not relevant at all for right-branching sentences.
We revisit this issue in the discussion part of this section.

\paragraph{Plotting trajectories}

\begin{figure}[hbtp]
    \begin{subfigure}{0.49\linewidth}
    \centering
        \includegraphics[width=\linewidth, trim={2cm, 0cm, 2cm, 3cm}]{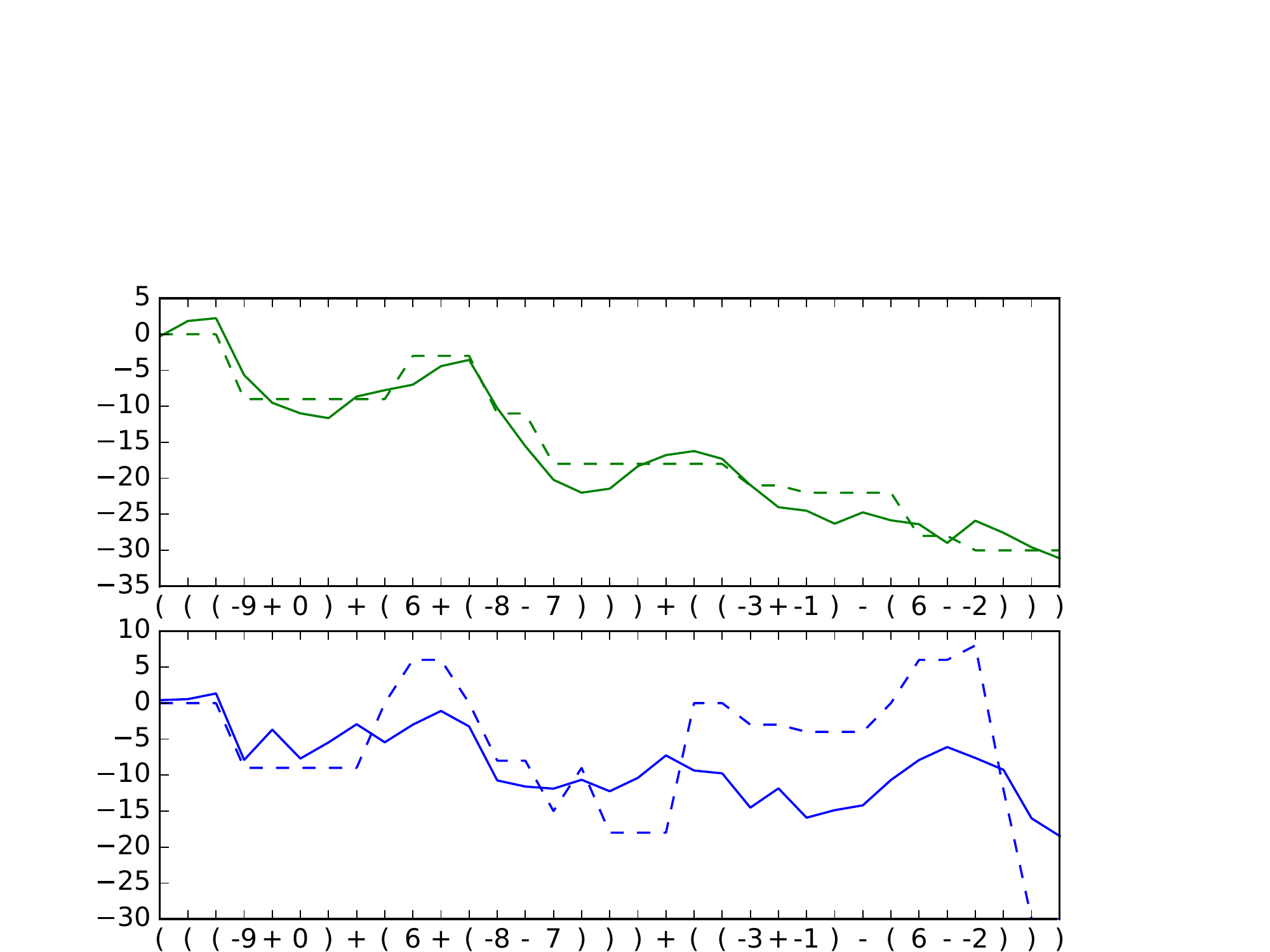}
    \end{subfigure}
    \centering
    \begin{subfigure}{0.49\linewidth}
        \includegraphics[width=\linewidth, trim={2cm, 2cm, 2cm, 3cm}]{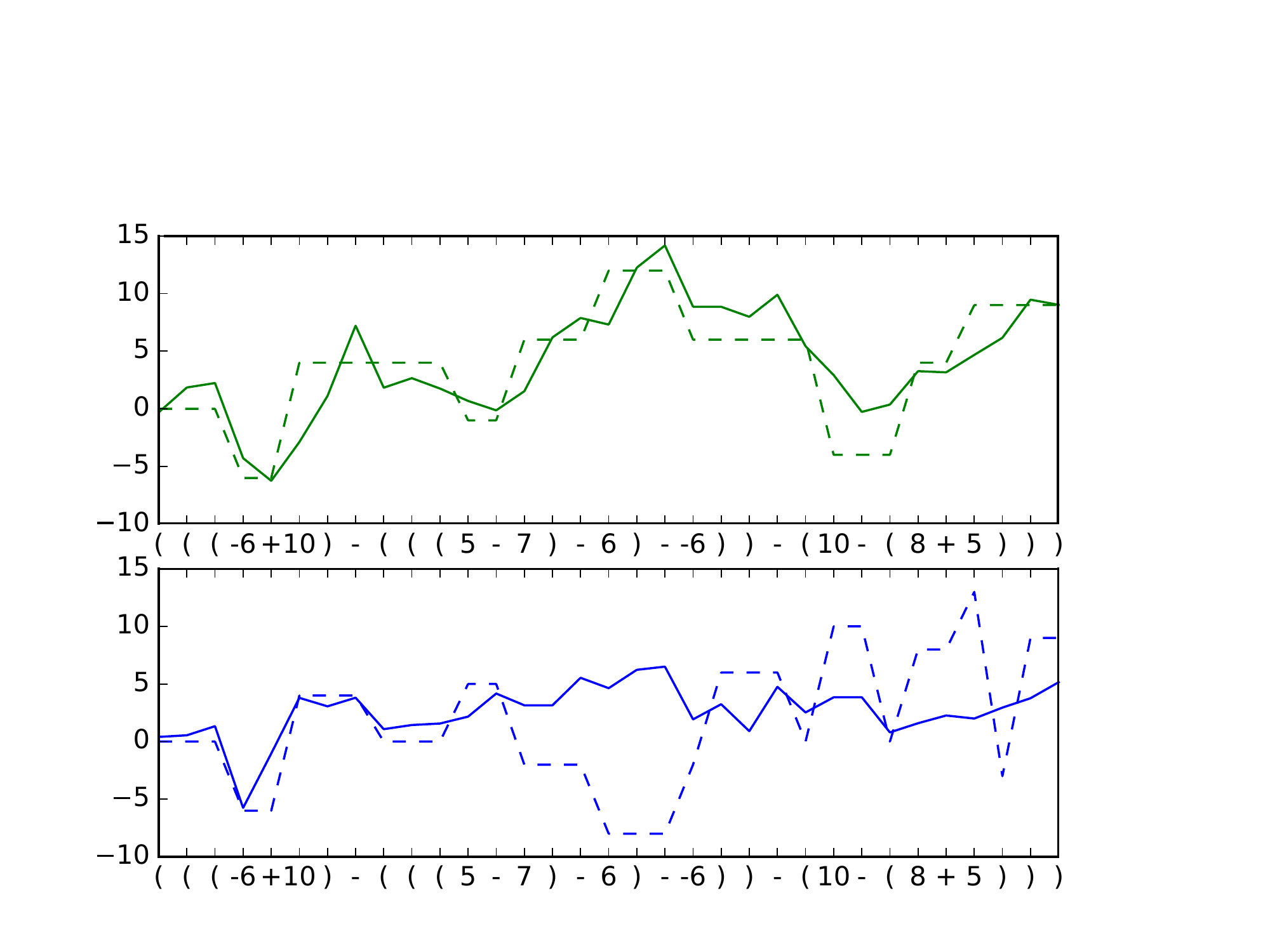}
    \end{subfigure}
\caption{Trajectories of the cumulative (green, upper) and recursive (blue, lower) classifier, along with their targets trajectories for the \texttt{result} values. Dashed lines show target trajectories.}\label{fig:trajectories2}
\end{figure}

We can not only use diagnostic classifiers to evaluate the overall match with a specific hypotheses, we can also track the fit of our predictions over time, by comparing the trajectories of predicted variables with the trajectories of observed variables while the networks process different sentences. 
In Figure~\ref{fig:trajectories2}, the predictions of the diagnostic classifiers on two randomly picked $\mathbf{L9}$ sentences are depicted, along with their target trajectories as defined by the hypotheses.
These trajectories confirm that the curve representing the cumulative strategy is much better predicted than the recursive one.
A correlation test over 5000 $\mathbf{L9}$ sentences shows the same trend: Pearson's $r$ = 0.52 and 0.95 for recursive and cumulative, respectively.
Figure~\ref{fig:results_c} shows the trajectory correlations for test sentences of different lengths.

However, we also observe an important qualitative difference between the diagnostic classifier trajectories and the target values: the diagnostic classifier trajectories are smooth, changing value at every point in time, whereas the target trajectories are jumpy and often stay on the same value for longer time spans.
This once more indicates that a refinement of the symbolic cumulative hypothesis, in which information is integrated more gradually, would be more suitable for a network like this.

\subsubsection{Discussion}

Using diagnostic classifiers, we have been able to analyse the internal dynamics of the GRU network in more detail than ever before. 
This allows us to conclude that the GRU is achieving its suprisingly good accuracy and generalisation behaviour by following a strategy that roughly approximates the symbolic `cumulative strategy'.
At the same time, it also provides us with evidence that the network's strategy does deviate from the cumulative strategy in details, and raises new questions.

\paragraph{Left- and right-branching expressions}
One of such questions concerns the difference between left- and right-branching expressions, and the fact that the GRU's behavior doesn't match the cumulative strategy's predictions very well for right-branching expressions. 
Understanding the reasons for this mismatch requires more research into the type of structures and operators that are difficult for the network or a more elaborate study of the trajectories of the trained diagnostic classifier for left- and right-branching sentences.

\paragraph{Diagnosing gates} Another open question concerns the exact role of the gating mechanism in computing the outcomes of arithmetic expressions. While we in this study focused on understanding the information in de hidden layer, diagnostic classifiers can certainly also be used to diagnose the role of \emph{gates}. Their use there does requires some extra care: where the hidden layer is directly connected to the output layer and it thus is reasonable to assume that computed information can be be predicted with a linear classifier, the information stored in the gates is used by the network only after passing through a multiplication and a non-linearity.
This suggests that a diagnostic classifier trained to diagnose gates should make use of a more complex architecture that can handle non-linearities, but this bears the risk that the classifier infers features that are not actually used by the network, but can be inferred as an artefact of the projection of the low dimensional word embeddings in a high dimensional space.

\paragraph{More realistic tasks} 
We have demonstrated that diagnostic classifiers can successfully extract interesting properties of low dimensional networks trained on an idealised task that would have been difficult to find otherwise. 
However, we have not yet experimented with deeper and higher dimensional networks and real world tasks. 
An exciting next step is using diagnostic classification for models trained on tasks such as language modelling, machine translation or semantic role labelling. 
Arguably, diagnostic classification could prove very useful for understanding the high dimensional networks typically used in these domains, for which visualisation is even more cumbersome.

%%%%%%%%%%%%%%%%%%%%%%%%%%%%%%%%%%%%%%%%%%%%%%%%%%%%%%%%%%%%%%%%%%%%%%%%%%%%%%%%
% Conclusions

\section{Conclusions}\label{sec:conclusion}

In this paper we studied how recursive and recurrent neural network process hierarchical structures, using an arithmetic language as a convenient, idealised task with unambiguous syntax and semantics and a limited vocabulary. 
We showed that recursive neural networks can learn to compute the meaning of arithmetic expressions and readily generalise to expressions longer than any seen in training. 
Learning was even successful when the representations were limited to two dimensions, allowing a geometric analysis of the solution found by the network.
Understanding the organisation of the word embedding space and the nature of the projections encoded in the learned composition function, combined with vector addition and squashing, gave us a complete picture of the compositional semantics that the network has learned.
This made clear that the network has found a near perfect approximation of a principled, recursive solution of the arithmetic semantics.

Still, the recursive neural network is a hybrid neural-symbolic architecture, as the network architecture directly reflects the tree structure of the expressions it processes, and this architecture is built up anew for each new expression using a symbolic control structure. 
This limits both the computational efficiency and the usefulness of this model for understanding how the human brain might process hierarchical structure. In this paper we therefore also analyse two recurrent neural network architectures: the classical Simple Recurrent Network and the recently popular Gated Recurrent Units. As it turns out, the latter can also learn to compute the meaning of arithmetic expressions and generalise to longer expressions than seen in training.
Understanding how this network solves the task, however, is more difficult due to its higher dimensionality, recurrent connections and gating mechanism.  To better understand what the network is doing we therefore developed an approach based on training diagnostic classifiers on the internal representations.

The qualitative and quantitative analyses of the results of a diagnostic classifier allow us to draw conclusions about possible strategies the network might be following.
In particular, we find that the successful networks follow a strategy very similar to our hypothesised symbolic `cumulative strategy'. From this we learn something about how neural networks may process languages with a hierarchical compositional semantics and, perhaps more importantly, also provide an example of how we can \textit{open the black box} of the many successful deep learning models in natural language processing (and other domains) when visualisation alone is not sufficient. 

\section*{Acknowledgements}

This paper is an extended version of the work of \citeA{veldhoen2016diagnostic}, presented at a workshop at NIPS 2016. 
DH and WZ are funded by the Netherlands Organization for Scientific Research (NWO), through a Gravitation Grant 024.001.006 to the Language in Interaction Consortium.

%%%%%%%%%%%%%%%%%%%%%%%%%%%%%%%%%%%%%%%%%%%%%%%%%%%%%%%%%%%%%%%%%%%%%%%%%%%%%%%%
% References

\bibliography{refs}
\bibliographystyle{theapa}

\end{document}